\documentclass[journal]{IEEEtran}
\usepackage{amsmath,amsfonts, amssymb}
\usepackage{soul}
\usepackage{acronym}
\usepackage{algorithmic}
\usepackage{algorithm}
\usepackage{array}
\usepackage[caption=false,font=normalsize,labelfont=sf,textfont=sf]{subfig}
\usepackage{textcomp}
\usepackage{stfloats}
\usepackage{verbatim}
\usepackage{cite}
\usepackage{bbm}
\usepackage{color}
\usepackage{tabularray}
\usepackage{graphicx}
\usepackage{orcidlink}
\usepackage{xcolor}
\usepackage{enumitem}
\newcommand{\footlink}[2]{#2 (\url{#1})}
\PassOptionsToPackage{hyphens}{url}\usepackage{hyperref}
\usepackage[normalem]{ulem}

\definecolor{Alto}{rgb}{0.93,0.93,0.93}
\definecolor{Camarone}{rgb}{0,0.392,0}
\definecolor{JapaneseLaurel}{rgb}{0,0.627,0}
\definecolor{DeepFir}{rgb}{0,0.235,0}
\definecolor{ChelseaGem}{rgb}{0.588,0.392,0}
\definecolor{PirateGold}{rgb}{0.666,0.45,0}
\definecolor{YellowOrange}{rgb}{1,0.705,0.196}
\definecolor{CaribbeanGreen}{rgb}{0,0.862,0.509}
\definecolor{LaserLemon}{rgb}{1,1,0.392}
\definecolor{MilkPunch}{rgb}{1,0.96,0.843}
\definecolor{MilanoRed}{rgb}{0.764,0.078,0}
\definecolor{DodgerBlue}{rgb}{0.356,0.584,1}
\definecolor{ScienceBlue}{rgb}{0,0.274,0.784}

\begin{document}
\title{Semantics-Aware Hierarchical Consensus Learning for Remote Sensing Image Classification}

\author{Giulio~Weikmann~\orcidlink{0000-0003-3762-8929},~\IEEEmembership{Member,~IEEE},
        Gianmarco~Perantoni~\orcidlink{0000-0002-1146-1035},~\IEEEmembership{Member,~IEEE},
        and~Lorenzo~Bruzzone~\orcidlink{0000-0002-6036-459X},~\IEEEmembership{Fellow,~IEEE}
\thanks{This work was supported in part by the European Space Agency (ESA) Climate Change Initiative (CCI+) New Essential Climate Variables (ECVs) High Resolution Land Cover (HRLC), under Contract 4000125259718/I-NB, and in part by the Italian Space Agency (ASI) and the Ministry of University and Research (MUR) under Contract 2024-5-E.0 - CUP n. I53D24000060005, by the Project ``Space It Up!''. \textit{(Corresponding author:
Lorenzo Bruzzone.)}}
\thanks{Giulio Weikmann, Gianmarco Perantoni and Lorenzo Bruzzone are with the Department of Information Engineering and Computer Science, University of Trento, 38123 Trento, Italy (e-mail: giulio.weikmann@unitn.it; gianmarco.perantoni@unitn.it; lorenzo.bruzzone@unitn.it).}
}

\markboth{Accepted for publication in IEEE Transactions on Geoscience and Remote Sensing}%
{Shell \MakeLowercase{\textit{et al.}}: A Sample Article Using IEEEtran.cls for IEEE Journals}

\maketitle

\begin{abstract}
Deep learning has become increasingly important in remote sensing image classification due to its ability to extract semantic information from complex data. Classification tasks often include predefined label hierarchies that represent the semantic relationships among classes. However, these hierarchies are frequently overlooked, and most approaches focus only on fine-grained classification schemes. In this paper, we present a novel Semantics-Aware Hierarchical Consensus (SAHC) approach that integrates hierarchical-level-specific classification heads within a deep network architecture and combines their output through cross-level probability projectors. Direct and projected predictions are fused into a geometric consensus distribution, which is used for self-consistent training and optional hierarchy-aware inference. This mechanism acts as a geometric ensemble that leverages the inherent structure of the hierarchical classification task. The projectors are initialized from the \textit{user-defined} taxonomy (\textit{i.e.} the hierarchical label structure), and can be adaptively refined during optimization.
The proposed SAHC method is evaluated on two benchmark datasets with different degrees of hierarchical complexity on different tasks, considering varying spectral and spatial resolutions. Experimental results show both the effectiveness of the proposed approach in guiding network learning and the robustness of the hierarchical consensus for remote sensing image classification tasks. The source code is available at \href{https://github.com/rslab-unitrento/sahc}{https://github.com/rslab-unitrento/sahc}.
\end{abstract}


\newacro{ALGS}{Adaptive Label Granularity Selection}
\newacro{CCE}{categorical cross-entropy}
\newacro{CCI}{Climate Change Initiative}
\newacro{CLC}{CORINE Land Cover}
\newacro{CNN}{Convolutional Neural Network}
\newacro{ConvLSTM}{Convolutional Long Short-Term Memory}
\newacro{CORINE}{COoRdination of INformation on the Environment}
\newacro{CV}{Computer Vision}
\newacro{DL}{Deep Learning}
\newacro{ELU}{Emilia Land Use}
\newacro{EO}{Earth Observation}
\newacro{ESA}{European Space Agency}
\newacro{HRN}{Hierarchical Residual Network}
\newacro{mF1}{mean F1 Score}
\newacro{MSE}{Mean Squared Error}
\newacro{FAO}{Food and Agriculture Organization}
\newacro{FCL}{Fully Connected Layer}
\newacro{FM}{Foundation Model}
\newacro{GSB}{Granularity-Specific Block}
\newacro{HAF}{Hierarchy Aware Features}
\newacro{HCL}{Hierarchical Contrastive Learning}
\newacro{HC}{Hierarchical Consensus}
\newacro{HRLC}{High Resolution Land Cover}
\newacro{HXE}{hierarchical cross-entropy}
\newacro{mIoU}{mean Intersection over Union}
\newacro{MGM}{multi-granularity module}
\newacro{MGC}{Multi-granularity Classification}
\newacro{JSD}{Jensen-Shannon Divergence}
\newacro{KLD}{Kullback–Leibler divergence}
\newacro{L2A}{Level-2A}
\newacro{LCCS}{Land Cover Classification System}
\newacro{LC}{Land-Cover}
\newacro{LSE}{log-sum-exp}
\newacro{LSTM}{Long Short-Term Memory}
\newacro{LULC}{Land-Use and Land-Cover}
\newacro{LU}{Land-Use}
\newacro{MGRS}{Military Grid Reference System}
\newacro{ML}{Maximum Likelihood}
\newacro{MS}{multispectral}
\newacro{NLP}{Natural Language Processing}
\newacro{OA}{Overall Accuracy}
\newacro{RF}{Random Forest}
\newacro{RNG}{random number generator}
\newacro{RS}{Remote Sensing}
\newacro{ResNet}{Residual neural network}
\newacro{SAHC}{Semantics-Aware Hierarchical Consensus}
\newacro{SAR}{Synthetic Aperture Radar}
\newacro{SITS}{Satellite Image Time Series}
\newacro{S1}{Sentinel-1}
\newacro{S2}{Sentinel-2}
\newacro{SVM}{Support Vector Machine}
\newacro{SWA}{Stochastic Weight Averaging}
\newacro{TPE}{Tree-structured Parzen Estimator}
\newacro{TS}{time series}
\newacro{TV}{Total Variation}
\newacro{ViT}{Vision Transformer}
\newacro{VHR}{Very-High Resolution}

\section{Introduction}\label{introduction}
\begin{figure}
  \includegraphics[width=\columnwidth]{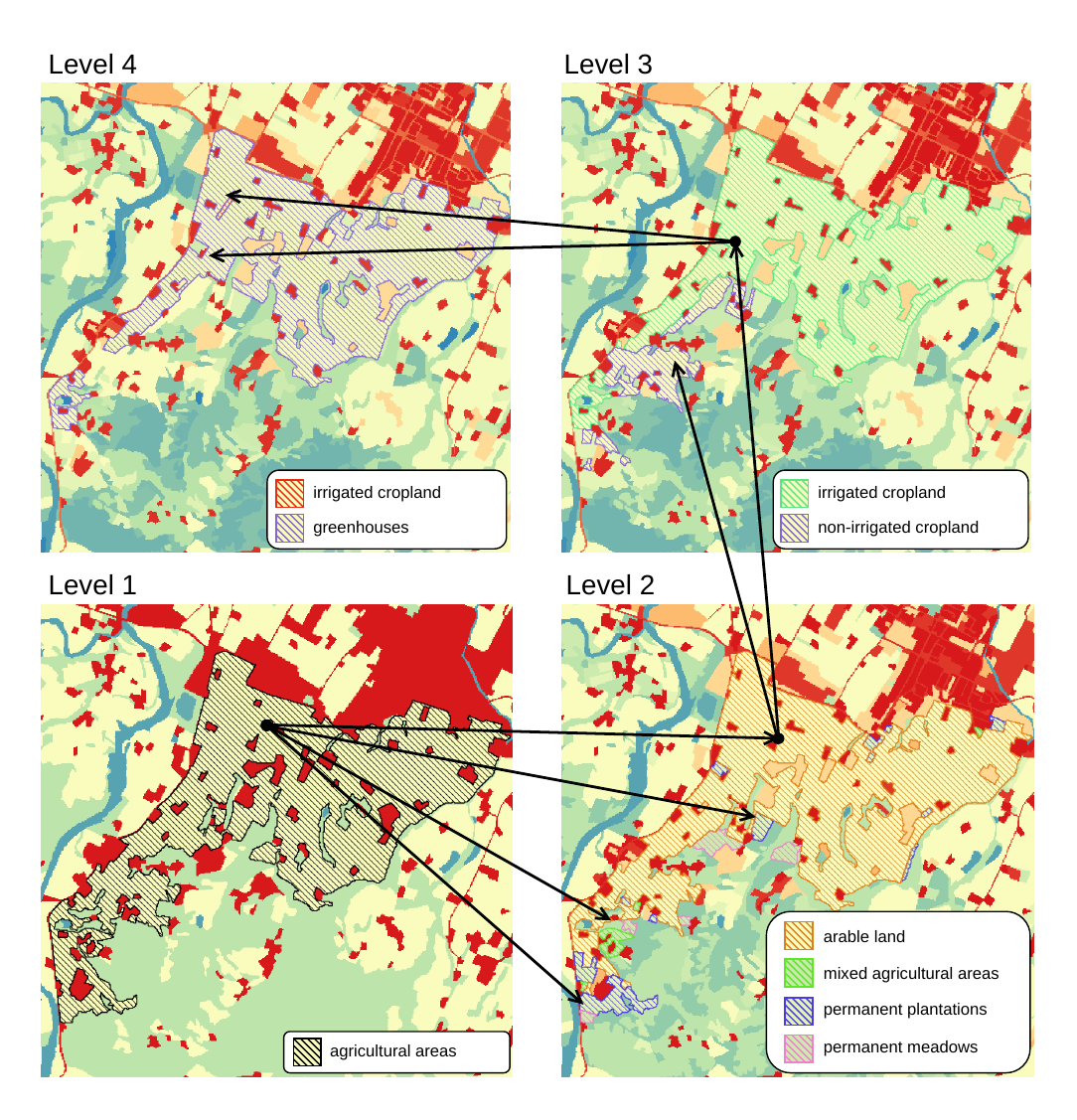}
  \caption{Hierarchical classification task example. Level 1 represents the coarsest class definition, where each class is progressively broken down into its fine-grained representations.}
  \label{fig:intro_example}
\end{figure}

\IEEEPARstart{I}{mage} classification and semantic segmentation are central tasks in \ac{RS} image processing. In recent years, the proliferation of \ac{RS} data and open distribution policies have significantly increased the availability of \ac{LULC} products, along with their corresponding classification schemes \cite{10.3389/frsen.2020.605220}. These products play a critical role in various applications and are widely used by government agencies, municipalities, and ministries. However, global and regional \ac{LULC} maps often differ in terms of resolution and are affected by semantic inconsistencies in class definitions and production algorithms \cite{wang2023review}. While \ac{DL} methods have revolutionized \ac{LULC} mapping by capturing complex spatial and spectral dependencies, standard approaches typically treat classification as a flat, single-level task. 
In practice, \ac{LC} classes are naturally organized into hierarchical label structures that reflect intrinsic levels of semantic granularity. This multi-level hierarchy is a property of the physical landscape. For instance, within a single dataset, a specific target might be unambiguously identified at a coarse-grained level (\textit{e.g.}, forest), while its finer-grained classification (\textit{e.g.}, deciduous versus coniferous) remains uncertain due to inherent label ambiguity, complex scene compositions, or the physical limits of the sensor. Neglecting this inherent hierarchical structure discards valuable semantic information and can lead to suboptimal or inconsistent predictions, particularly when dealing with label ambiguity or complex \ac{RS} scenes. An example of this hierarchical problem definition can be seen in Fig. \ref{fig:intro_example}.

In the literature, most of the works related to \ac{RS} image classification focus on single-level approaches, discarding the hierarchical information contained in the data. Hierarchical classification schemes, such as the \ac{CLC} maps, have mostly been adopted in object-based image classification methods based on \ac{VHR} satellite images \cite{BLASCHKE20102}, often considering a limited number of classes to simplify the classification task. To validate the hierarchical structure, few works analyzed the inter-level relationships, considering decision trees in an object-based classification scenarios \cite{ijgi7100408}, or validation approaches based on intercomparison of different maps \cite{perez2013incorporating}. Instead, the wide majority of works in the \ac{RS} domain focus on single-level approaches, discarding the hierarchical information contained in the data. While these methods show promising results at individual hierarchical levels, they fail to incorporate the hierarchical structure of the data into the decision-making process, overlooking important information included in the source data.

Only recently, hierarchical and multi-granularity classification methods have emerged to address these challenges in \ac{RS}. These methods leverage the hierarchical structure to encode semantic relationships among labels, enabling knowledge transfer across different levels of granularity. Recent studies have demonstrated the benefits of adaptive label granularity selection to balance classification accuracy and specificity \cite{chen2025adaptive}, as well as the integration of parent-child relationships and sibling mutual exclusivity to improve hierarchical representation learning \cite{chen2025hierarchical}. Despite these advancements, integrating hierarchical information effectively into the learning process remains an open challenge. Many existing methods rely on complex, task-specific architectures or rigid rule-based post-processing steps, which limit their flexibility across different \ac{DL} backbones and fail to explicitly enforce probability consistency across different hierarchical levels in a generalized manner.

To address these limitations, we propose a novel and modular \ac{SAHC} method. SAHC integrates hierarchical-level-specific classification heads within standard \ac{DL} architectures, projects multi-level predictions to a common hierarchical level, and combines them through a consensus distribution. The corresponding projectors can be derived from the \textit{user-defined} taxonomy (\textit{i.e.}, the hierarchical label structure), while trainable hierarchy projectors provide an optional adaptive refinement around this semantic prior.

The key contributions of this work can be summarized as follows:

\begin{enumerate}
    \item We introduce a Semantics-Aware Hierarchical Consensus (SAHC) framework that treats the predictions produced at different hierarchical levels as a committee of probabilistic estimates. The predictions are projected to a common target level and fused into a consensus distribution.
    \item We propose a self-consistent learning strategy that aligns direct and projected predictions with the consensus during training. The same mechanism supports hierarchy-aware consensus prediction at inference, while direct-head predictions remain available when fine-level discrimination is the primary objective.
    \item We introduce taxonomy-initialized trainable projection matrices as an adaptive parameterization of the cross-level projectors. This parameterization enables \textit{data-driven} calibration of cross-level compatibility while preserving the \textit{user-defined} taxonomy as a semantic prior.
    \item We demonstrate the effectiveness and robustness of the proposed approach on two \ac{RS} datasets with different hierarchy complexity, spatial resolution, and classification settings.
\end{enumerate}

The remainder of this paper is organized as follows. Section \ref{relatedworks} reviews related works on hierarchical semantic segmentation. Section \ref{proposed} details the proposed hierarchical consensus approach. The datasets and experimental setup are described in Section \ref{dataset}, while Section \ref{results} presents and discusses the experimental results. Finally, Section \ref{conclusion} concludes the paper and outlines future research directions.

\section{Related Work}\label{relatedworks}

Image classification and semantic segmentation are widely studied tasks in \ac{RS} image processing, with \ac{DL} models achieving state-of-the-art performance across various domains. Standard approaches typically treat classification as a ``flat", single-granularity problem, predicting probabilities over a set of mutually exclusive classes without explicitly considering potential semantic relationships or hierarchical structures among them. However, many real-world \ac{RS} datasets possess inherent taxonomies where classes can be naturally organized into hierarchies (\textit{e.g.}, standardized land-cover systems, target recognition taxonomies with super- and sub-categories). Ignoring this structure can lead to suboptimal performance, as standard loss functions treat all misclassifications equally. Without an explicit hierarchical representation, models fail to differentiate between errors involving closely related classes and those involving semantically distant ones.

Leveraging hierarchical label structures during model training and inference has emerged as a promising direction to improve accuracy and semantic understanding. Early work demonstrated that standard \acp{CNN} implicitly capture some hierarchical information, with shallower layers grouping high-level categories and deeper layers specializing in finer distinctions \cite{bilal_jourabloo_ye_liu_ren_2018}. Building on this, researchers have explored explicit methods to incorporate hierarchy. Typically, this is achieved by mapping the label hierarchy to specific network architectures or by introducing loss functions that handle the correlation between different granularity tasks \cite{CERRI201439, 7849143, giunchiglia2020coherent}. For instance, several methods propose level-wise hierarchical cross-entropy losses based on conditional probabilities, or introduce custom features to minimize the divergence between aggregated predictions at different hierarchical levels, aiming to reduce the severity of misclassification. Bertinetto \textit{et al.} \cite{bertinetto2020making} proposed a level-wise hierarchical cross-entropy loss based on conditional probabilities within the hierarchy and used mapping functions to embed class relationships. Similarly, Garg \textit{et al.} \cite{garg2022learning} introduced the \ac{HAF}, incorporating multiple losses, including minimizing the \ac{JSD} between aggregated predictions at different hierarchical levels and enforcing geometric consistency in the feature space, aiming to reduce the severity of misclassifications. In \cite{deng2014large}, the authors proposed hierarchy and exclusion graphs to model semantic relations between two labels assigned to the same objects. They considered exclusions, overlaps, and subsumptions, and provided a detailed theoretical analysis of the structure of the hierarchical problem. Building on this work, a combinatorial loss is proposed by Chen \textit{et al.} \cite{chen2022label}, which addresses the scenario where samples might be labelled at different levels of granularity. The loss is designed to maximize the marginal probability of the observed ground truth label by aggregating information from related labels in the hierarchy. The authors introduced a \ac{HRN} architecture, where features from parent levels act as residual connections to inform child-level features, explicitly facilitating hierarchical feature interaction. Following this work, Zhou \textit{et al.} \cite{Zhou_Wei_Zhang_Qi_Yang_Li_2023} proposed a multi-granularity archaeological dating method that depends on an explicit knowledge-guided relation graph, employing a focal-type probability classification loss to mine the relationship between specific object attributes and their broader eras. Their network leverages a bidirectional multi-granularity module with gradient truncated addition to explicitly fuse coarse- and fine-grained features. Similarly, Cai \textit{et al.} \cite{hicervix} introduced HierSwin, a hierarchical vision transformer, which explicitly models the semantic correlations embedded within a predefined three-level hierarchical tree to gradually improve accuracy from fine- to coarse-grained classifications.

More recently, some works addressed the \ac{MGC} in \ac{RS} to handle targets that are discernible only at specific variable semantic levels due to differences in spatial resolution, sensor modalities, and data quality. Chen \textit{et al.} \cite{chen2025adaptive} introduced an accuracy-specificity curve to adaptively select task-dependent prediction thresholds. Their approach also employs a penalty term based on the global semantic distance across the label tree to adjust class logits. Addressing a similar challenge in ship classification, Chen \textit{et al.} \cite{chen2025hierarchical} proposed a hierarchical contrastive learning framework. By explicitly incorporating parent-child dependencies and mutual exclusivity among sibling categories, their method aligns class-specific representations with semantic relationships and enforces probability coherence across levels through a consistency loss.

Despite these significant advancements, many existing multi-granularity and hierarchical learning methods still rely on the assumption that a predefined and accurate hierarchy is available, where no hierarchy violation is possible. This requires rigid post-processing rules, fixed decision thresholds, or highly specific non-modular architectures. Furthermore, many existing techniques require explicit guidance or architectural modifications tailored to a specific fixed hierarchy \cite{9311828}, potentially limiting their flexibility and ability to adapt if the underlying hierarchical understanding evolves or is uncertain. For example, some approaches decompose the problem into sequential decisions or binary classifications guided by the known structure \cite{demirkan2020hierarchical,wasniewski2022can}, rather than allowing the model to learn or refine hierarchical relationships implicitly from the data. 

The challenge remains to exploit predefined hierarchical semantics through a modular probability-driven mechanism, while allowing limited \textit{data-driven} calibration of cross-level relationships around the taxonomy prior.

\section{Proposed Methodology}\label{proposed}
In this work, we address the above-mentioned challenges by proposing a novel deep learning framework for hierarchy-aware image classification. Our approach models the hierarchical semantic relationships within image data by integrating hierarchical loss components with a robust consensus mechanism. 

The main methodological component of \ac{SAHC} is an all-level hierarchical consensus approach. For each target hierarchical level, the predictions produced by all hierarchical levels are projected into corresponding target label space through cross-level projectors. These projectors can be directly derived from the \textit{user-defined} taxonomy and, in the adaptive implementation, are initialized from this taxonomy and jointly optimized with the network. Then, the direct prediction of the target-level head and the projected predictions from all remaining levels are fused into a consensus probability distribution. \ac{SAHC} jointly learns level-specific classifiers and taxonomy-anchored probabilistic mappings that align their predictions across the hierarchy.

The proposed \ac{SAHC} differs from other consistency-based approaches in the way hierarchical agreement is modeled. Previous methods generally enforce local parent-child coherence through fixed child-to-parent probability aggregation, pairwise consistency constraints, or semantic-distance-based penalties. In contrast, \ac{SAHC} projects the predictions produced at all hierarchical levels into a common target label space and jointly combines them with the direct target-level prediction through a consensus distribution. This consensus is used both as a regularization target during training and as an optional hierarchy-aware prediction at inference.

\subsection{Hierarchical Problem Formulation}
Let us denote column vectors by lowercase bold letters (\textit{e.g.}, $\mathbf{v}$) and matrices by uppercase bold letters (\textit{e.g.}, $\mathbf{B}$). The $i$th component of a vector $\mathbf{v}$ is denoted by $v_{i}$, while the $i$th row and $j$th column of a matrix $\mathbf{B}$ are denoted by $\mathbf{B}_{i,\cdot}$ and $\mathbf{B}_{\cdot,j}$, respectively. Then, let us consider a user-defined hierarchy label tree with $H$ levels, where the leaves represent the finest level of detail. We denote each hierarchical level as $h\in\mathcal{H}=[1,2,\dots,H]$, ranging from the coarsest level $h=1$ to the finest one $h=H$. The root node $h=0$ is not considered, as it represents the trivial case in which all classes are mapped into a single class. Let $\mathbf{x}\in{\mathbb{R}^{b}}$ be the input vector, where $b$ is the number of input features. Its associated labels are $y^{h}\in \Omega^{h}$, where $\Omega^{h}=\{\omega^{h}_{1},\dots,\omega^{h}_{|\Omega^{h}|}\}$ is the set of labels at hierarchical level $h$, with $|\Omega^{h}|$ number of labels. For every $h_{1}<h_{2}$, we have that each fine-grained class $\omega_{j}^{h_2}$ is mapped to a coarse-grained class $\omega_{i}^{h_1}$, and $|\Omega^{h_1}|\le|\Omega^{h_2}|$. We define this mapping relationship by $\omega_{j}^{h_2}\implies\omega_{i}^{h_1}$. Let $\mathcal{D}=\{(\mathbf{x}_{n},Y_{n})|n=1,\dots,N\}$ be the training set, where for each input feature vector $\mathbf{x}_{n}$, we have an associated set $Y_{n}=\{y_{n}^{1},\dots,y_{n}^{h},\dots,y_{n}^{H}\}$, which denotes the ground truth labels at all $H$ hierarchical levels, for a total number of $N$ training samples.
The labels in the dataset satisfy a nested label structure, where each class at a finer level is contained within a class at a coarser level.

\subsection{Fully-Supervised Multi-Level Learning}
\begin{figure*}
    \includegraphics[width=\textwidth]{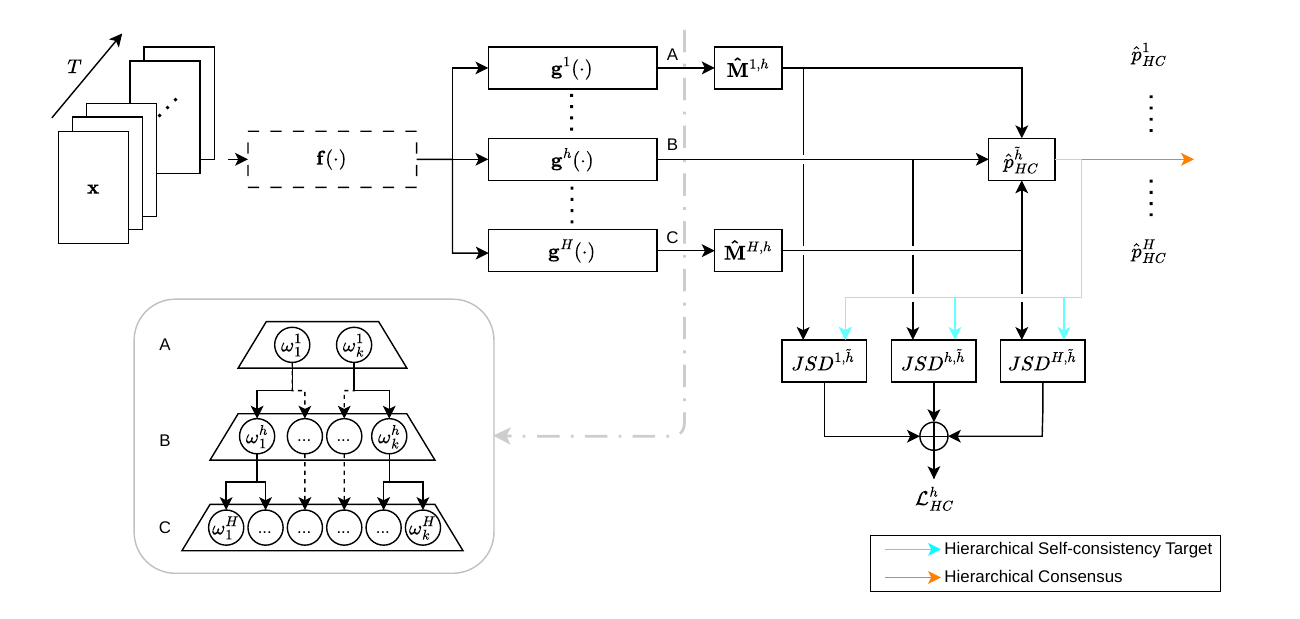}
    \caption{Illustration of the proposed hierarchical architecture with the hierarchical consensus calculated on $h=\tilde{h}$. Heads tailored to different granularity levels are reprojected on the target level and geometrically averaged to extract the consensus probability $\hat{p}^{\tilde{h}}_{HC}$. 
    This consensus probability is used to calculate the hierarchical self-consistency loss $\mathcal{L}^{h}_{HC}$. Optionally, this consensus probability can be used to generate a robust hierarchy-aware prediction during inference.}
    \label{fig:hierarchic_repr}
\end{figure*}

The proposed methodology can be applied to any backbone network by simply introducing the required additional hierarchical classification modules. We refer to each of the hierarchical classification modules $\mathbf{g}^{h}:{\mathbb{R}^{d}\rightarrow\mathbb{R}^{|\Omega^{h}|}}$ on top of a given backbone network as the composition $\mathbf{g}^{h}\circ\mathbf{f}:{\mathbb{R}^{b}\rightarrow\mathbb{R}^{|\Omega^{h}|}}$, where $\mathbf{f}:{\mathbb{R}^{b}\rightarrow\mathbb{R}^{d}}$ is the backbone feature extractor. Fig. \ref{fig:hierarchic_repr} provides a high-level overview of the considered multi-level classification approach.

Each classifier $\mathbf{g}^{h}$ estimates class-posterior logits. Let $\hat{\mathbf{p}}^{h}(\mathbf{x})\in\mathbb{R}^{|\Omega^{h}|}$ be the vector of estimated class-posterior probabilities at level $h$ given $\mathbf{x}$, \textit{i.e.}, $\hat{\mathbf{p}}^{h}(\mathbf{x}) = [\hat{P}(\omega_{1}^{h}|\mathbf{x}),\dots,\hat{P}(\omega_{|\Omega^{h}|}^{h}|\mathbf{x})]^{\mathsf{T}}$, where $\mathsf{T}$ refers to the transpose operator. Then, the class-posterior probabilities $\hat{\mathbf{p}}^{h}$ for the input sample $\mathbf{x}$ at level $h$ can be retrieved as follows:
\begin{equation}
    \hat{\mathbf{p}}^{h}\left(\mathbf{x}\right) = \boldsymbol{\sigma}\left[\mathbf{g}^{h}\left(\mathbf{f}\left(\mathbf{x}\right)\right)\right],
\end{equation}
where $\boldsymbol{\sigma}:{\mathbb{R}^{d}\rightarrow[0,1]^{d}}$ is the \textit{softmax} operator, with $\sigma_{i}[\cdot]$ computed as follows:
\begin{equation}\label{eq:classifiers}
    \sigma_{i}\left[\mathbf{g}\right] = \frac{\exp{g_{i}}}{\sum_{j} \exp{g_{j}}}\in\left[0,1\right].
\end{equation}
The training of $\mathbf{g}^{h}$ is done by computing the \ac{CCE} loss as follows:
\begin{equation}
    \begin{split}
    \mathcal{L}_{CCE}^{h} &= -\frac{1}{N}\sum_{n=1}^{N}\sum_{i=1}^{|\Omega^{h}|} \mathbbm{1}\left[y^{h}_{n}=\omega^{h}_{i}\right]\log\hat{P}^{h}\left(\omega^{h}_{i}|\mathbf{x}_{n}\right)\\
    &= -\frac{1}{N}\sum_{n=1}^{N}\sum_{i=1}^{|\Omega^{h}|}\mathbbm{1}\left[y^{h}_{n}=\omega^{h}_{i}\right]\log\sigma_{i}\left[\mathbf{g}^{h}\left(\mathbf{f}\left(\mathbf{x}_{n}\right)\right)\right],
    \end{split}
\end{equation}
where $\mathbbm{1}\left[\cdot\right]$ serves as an indicator function and takes a value of one when the argument is true, else zero.
We apply the \ac{CCE} loss function to all hierarchical levels, obtaining the total \ac{CCE} loss $\mathcal{L}_{CCE}$ as the sum of the level-wise losses:
\begin{equation} \label{eq:cce}
    \mathcal{L}_{CCE} = \sum_{h=1}^{H} \lambda^{h}\mathcal{L}_{CCE}^{h}.
\end{equation}
As it has also been noted in previous works \cite{chang2021your}, adopting several \ac{CCE} loss functions for diverse classification levels may negatively affect performance at the most granular classification level. To mitigate this behavior, we introduced the weighting terms $\lambda=\{\lambda^{1},\dots,\lambda^{H}\}$ to limit the impact of coarser levels, while still accounting for the additional losses.

\subsection{Cross-Level Projectors}
\ac{SAHC} requires cross-level projection operators to map predictions between hierarchical levels before consensus formation. These projectors can be directly derived from the \textit{user-defined} taxonomy or adaptively parameterized. The implementation considered in this work uses trainable, taxonomy-initialized matrices to refine the projectors during optimization.

The \ac{CCE} loss functions at each hierarchical level lead to independent classifications, which are not constrained to follow the hierarchy definition formulated in the problem statement. To promote cross-level agreement, additional consensus predictions and self-consistent penalty terms are introduced. These penalty terms are defined through hierarchy projection matrices designed to map the taxonomy-specified relationships between the different layers. 

\begin{figure}
    \centering
    \includegraphics[width=0.8\columnwidth]{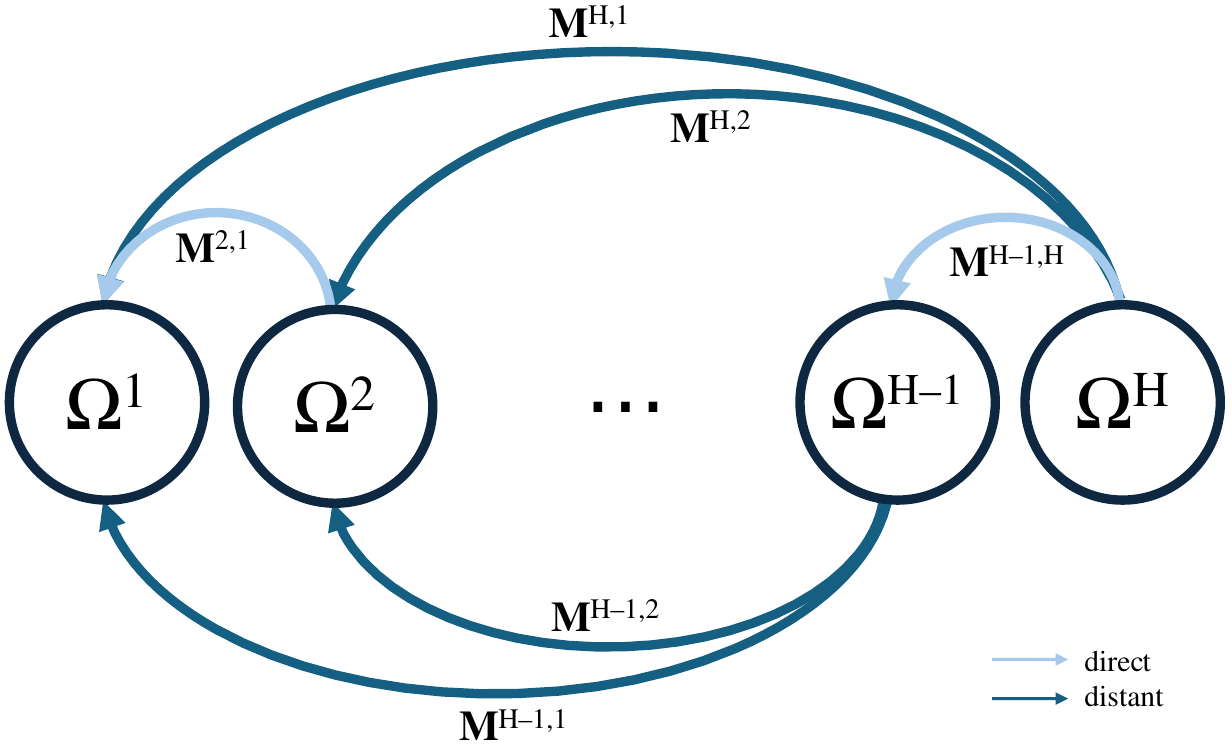}
    \caption{Illustration of both direct and distant hierarchical mappings, from fine-grained classes to coarse-grained classes.}
    \label{fig:transitions_repr}
\end{figure}

Let us consider for simplicity the case of $H=2$ levels, with $h_{1}$ representing the coarse-grained labels and $h_{2}$ the fine-grained labels, \textit{i.e.}, $h_{1}<h_{2}$. We refer to $\Omega^{h_1}$ as the set of the \textit{macroclasses} (\textit{i.e.,} the coarse-grained classes) and to $\Omega^{h_2}$ as the set of the \textit{microclasses} (\textit{i.e.}, the fine-grained classes). We can define the hierarchy indicator matrix $\mathbf{I}^{h_{2},h_{1}}\in\{0,1\}^{|\Omega^{h_{2}}|\times|\Omega^{h_{1}}|}$, which represents the \textit{user-defined} hierarchical relationship between level $h_2$ and $h_1$:

\begin{equation}
\mathbf{I}^{h_{2},h_{1}} = \begin{bmatrix}
I_{1,1} & I_{1,2} & \dots & I_{1,|\Omega^{1}|} \\
I_{2,1} & I_{2,2} & \dots & I_{2,|\Omega^{1}|} \\
\vdots & \vdots & \ddots & \vdots \\
I_{|\Omega^{2}|,1} & I_{|\Omega^{2}|,2} & \dots & I_{|\Omega^{2}|,|\Omega^{1}|}
\end{bmatrix}.
\end{equation}
This approach can be easily extended to a multi-level scenario, considering any given number $H$ of hierarchical levels. In this case, the total number of hierarchy matrices that can be determined is $((H-1)\cdot H)/2$, as also mappings of distant relationship levels can be defined (\textit{e.g.}, from the root of the label tree to the leaves, skipping intermediate levels) if needed. A generalized example is illustrated in Fig. \ref{fig:transitions_repr}.
In a strict hard-hierarchy, $I^{h_{2},h_{1}}_{i,j}$ assumes a value of one when the microclass $\omega_{i}^{h_{2}}$ is contained in the macroclass $\omega_{j}^{h_{1}}$ through the relationship ${\omega_{i}^{h_{2}}\implies\omega_{j}^{h_{1}}}$, and zero otherwise. However, this rigid formulation often fails to account for semantic overlaps in real-world hierarchies. Therefore, we distinguish between \textit{user-defined} aggregations, established a priori for a given task, and \textit{data-driven} adaptive projection weights, which calibrate dataset-specific associations and model uncertainties around the taxonomy prior. 
To combine the \textit{user-defined} prior with a \textit{data-driven} optimization, we parameterize the hierarchical relationships in the unconstrained logit space. Instead of strictly fixing the matrices to binary values or directly bounding them in $[0,1]$, we define learnable parameter matrices $\mathbf{W}^{h_{2},h_{1}}\in{\mathbb{R}}^{|\Omega^{h_2}|\times|\Omega^{h_1}|}$ used to model the log joint distribution matrices $\log{\mathbf{J}^{h_{2},h_{1}}}\in{\mathbb{R}}^{|\Omega^{h_2}|\times|\Omega^{h_1}|}$. $\mathbf{W}^{h_{2},h_{1}}$ are initialized based on the \textit{user-defined} indicator matrix $\mathbf{I}^{h_{2},h_{1}}$ to provide a stronger prior:
\begin{equation} \label{eq:wfromindicator}
W^{h_{2},h_{1}}_{i,j} =
    \begin{cases}
        \epsilon & \text{if } {I^{h{2},h_{1}}_{i,j} = 1} \\
        -\Delta + \epsilon & \text{if } {I^{h{2},h_{1}}_{i,j} = 0}
    \end{cases},
\end{equation}
where $\epsilon\sim\mathcal{N}(0,s^{2})$ is a small random element-wise noise added to break network symmetry while still maintaining the injected structural prior during initial phases of optimization. The scalar $\Delta$ defines the logit gap between \textit{user-defined} connections and unmapped relations. The two values act as positive and negative biases towards valid connections and unmapped relationships. The value of $\Delta$ is selected in the unconstrained logit space to encode a soft structural prior over the hierarchy and is used to provide small non-zero probability values to unmapped relations in the \textit{user-defined} taxonomy. Indeed, during the forward pass, these unconstrained parameters are dynamically mapped into valid hierarchy log joint distribution matrices $\log{\mathbf{J}^{h_{2},h_{1}}}$ via a log-softmax operation on the flattened parameter matrices $\mathbf{W}^{h_{2},h_{1}}$. Each term of the matrix $J^{h_{2},h_{1}}_{i,j}=P(\omega^{h_1}_{j},\omega^{h_2}_{i})$ models the probability of observing a relationship between microclass $\omega^{h_2}$ and macroclass $\omega^{h_1}$.
This constraint relaxation allows us to better combine the \textit{user-defined} aggregation with \textit{data-driven} calibration of cross-level compatibility, while retaining the initial \textit{user-defined} mappings as the structural anchor.

\subsection{Self-Consistent Hierarchical Consensus Learning}
Let us return to the general case of $H$ hierarchical levels, considering projections from source levels $h$ to a target hierarchical level $\tilde{h}$.
Each hierarchical level acts independently of the others. To ensure that the network is \textit{hierarchy-aware}, an additional constraint based on the prediction associated to the target hierarchical level and the projected predictions from the other levels is defined. In particular, we take inspiration from the disagreement-based active learning field \cite{Settles2012} to define the disagreement between the predictions calculated at each hierarchical level. For each level $\tilde{h}$, we consider a committee composed of all $H$ levels. All predictions and their projected versions can be viewed as ``soft'' votes, which define each committee member's confidence. Then, we introduce a self-consistent loss that penalizes the committee members' disagreement with the overall committee consensus, calibrating the mapping weights in $\mathbf{J}^{h,\tilde{h}}$ when adaptive projectors are used, and providing further supervision to the network based on the predictions at the different levels.

\subsubsection{Hierarchy-based Projected Predictions}
In order to generate the committee at each hierarchical level, the predictions first need to be projected to the other levels.
For simplicity, let us consider the projection from a source-level $h$ to a target level $\tilde{h}$. Let $\hat{\mathbf J}^{h,\tilde h}\in[0,1]^{|\Omega^h|\times|\Omega^{\tilde h}|}$, with $h,\tilde{h}\in\mathcal{H},h\neq\tilde{h}$, denote the joint distribution matrices.
From $\hat{\mathbf J}^{h,\tilde h}$, we derive the conditional projection matrix $\hat{\mathbf M}^{h,\tilde h}\in[0,1]^{|\Omega^h|\times|\Omega^{\tilde h}|}$ by row-normalization:
\begin{equation} \label{eq:mdefinition}
\hat M_{i,j}^{h,\tilde h}
=
\hat P(\omega_j^{\tilde h}\mid \omega_i^h)
=
\frac{
\hat J_{i,j}^{h,\tilde h}
}{
\sum_{k=1}^{|\Omega^{\tilde h}|}
\hat J_{i,k}^{h,\tilde h}
}.
\end{equation}
Therefore, the projected probability distribution $\hat P^{h\rightarrow\tilde h}(\omega_j^{\tilde h} | \mathbf{x})$ is obtained from the source-level posterior $\hat P^h(\omega_i^h | \mathbf{x})$ and the conditional projection probabilities $\hat P(\omega_j^{\tilde h} | \omega_i^h)$ as follows:
\begin{equation} \label{eq:projnovec}
    \hat{P}^{h\rightarrow \tilde{h}}{\left(\omega^{\tilde{h}}_{j}|\mathbf{x}\right)}=
    \sum_{i=1}^{|\Omega^{h}|}
    \hat{P}{\left(\omega^{\tilde{h}}_{j}|\omega^{h}_{i}\right)}
    \cdot 
    \hat{P}^{h}{\left(\omega^{h}_{i}|\mathbf{x}\right)},
\end{equation}
This corresponds to the input-independent hierarchy projection assumption $\hat P(\omega_j^{\tilde h} | \omega_i^h,\mathbf{x})\approx\hat P(\omega_j^{\tilde h} | \omega_i^h)$. The dependency on the input $\mathbf{x}$ is preserved through the source-level posterior $\hat P^h(\omega_i^h | \mathbf{x})$, while the conditional hierarchy transition is modeled as a global projection probability shared across the dataset.
In general vector form, we can write the class-posterior probability vector $\hat{\mathbf{p}}^{h\rightarrow\tilde{h}}{(\mathbf{x})}$ projected from level $h$ to level $\tilde{h}$ as follows:
\begin{equation} \label{eq:proj-nolog}
    \hat{\mathbf{p}}^{h\rightarrow\tilde{h}}{(\mathbf{x})} = (\hat{\mathbf{M}}^{h,\tilde{h}})^{\mathsf{T}} \cdot \hat{\mathbf{p}}^{h}{\left(\mathbf{x}\right)}.
\end{equation}
The row-normalization in \eqref{eq:mdefinition} constrains each row in $\hat{\mathbf M}^{h,\tilde h}$ to sum to one. Therefore, although the hierarchy parameters $\mathbf{W}^{h,\tilde h},h,\tilde{h}\in\mathcal{H},h\neq\tilde{h}$, are unconstrained in the logit space, the quantities used to project predictions across hierarchical levels are valid conditional probability distributions. 

To ensure numerical stability and maintain better control over gradient magnitudes, we consider the computation of all the losses directly in the log domain. First, we compute the class-posterior log probabilities of classification modules $\mathbf{g}^{h}$ from \eqref{eq:classifiers}:
\begin{equation}
    \log{\hat{P}^{h}}{\left(\omega_{i}^{h}|\mathbf{x}\right)} = g_{i}^{h}-\text{LSE}{\left(\mathbf{g}^{h}\right)},
\end{equation}
where $\text{LSE}:{\mathbb{R}^{|\Omega^{h}|}\rightarrow\mathbb{R}}$ is the \ac{LSE} operator:
\begin{equation}
    \text{LSE}\left(\mathbf{g}^{h}\right)=\log\sum_{i=1}^{|\Omega^{h}|}{\exp{\left(g_{i}^{h}\right)}}.
\end{equation}
Then, we can also define the projected predictions $\hat{P}^{h\rightarrow\tilde{h}}{(\omega_j^{\tilde{h}}|\mathbf{x})}$ from level $h$ to level $\tilde{h}$ in the log domain, starting from the definition in the natural domain. 
Applying the logarithm to both sides of \eqref{eq:projnovec}, we obtain the following:
\begin{align}
    \log{\left(\hat{P}^{h\rightarrow\tilde{h}}{\left(\omega^{\tilde{h}}_{j}|\mathbf{x}\right)}\right)} = \notag \\
    &\hspace{-7em} = \log{\left(\sum_{i}\hat{M}^{h,\tilde{h}}_{i,j}\cdot \hat{P}^{h}{\left(\omega^{h}_{i}|\mathbf{x}\right)}\right)} \notag\\
    &\hspace{-7em} = \log{\left(\sum_{i}\hat{M}^{h,\tilde{h}}_{i,j}\cdot \frac{\exp{g_{i}^{h}}}{\sum_{k} \exp{g_{k}^{h}}}\right)} \notag\\
    &\hspace{-7em} = \log{\left(\sum_{i}\exp\left({g_{i}^{h}+\log{\hat{M}^{h,\tilde{h}}_{i,j}}}\right)\right)} - \log{\sum_{k}\exp{g_{k}^{h}}}.
\end{align}
This can finally be rewritten as the combination of two \ac{LSE} operations, yielding the projected log-probability distribution at level $\tilde{h}$:

\begin{multline} \label{eq:proj-log}
    \log{\left(\hat{P}^{h\rightarrow\tilde{h}}\left(\omega_{j}^{\tilde{h}}|\mathbf{x}\right)\right)} = \\ = \text{LSE} \left(\mathbf{g}^{h} + \log{\left(\mathbf{\hat{M}}^{h,\tilde{h}}_{\cdot,j}\right)}\right) - \text{LSE}\left(\mathbf{g}^{h}\right).
\end{multline}
Hierarchy projection matrices can be defined directly in the log domain, \textit{i.e.}, we retrieve log joint probabilities $\log{\hat{\mathbf{J}}^{h,\tilde{h}}}\in\mathbb{R}^{|\Omega^{h}|\times|\Omega^{\tilde{h}}|}$ via learned parameter matrices $\mathbf{W}^{h,\tilde{h}}$:
\begin{equation}
        \log{\hat{\mathbf{J}}^{h,\tilde{h}}}=\mathbf{W}^{h,\tilde{h}}-\text{LSE}\left(Flatten\left(\mathbf{W}^{h,\tilde{h}} \right)\right)
\end{equation}
Then, we compute $\log{\hat{\mathbf{M}}^{h,\tilde{h}}}$ as follows:
\begin{equation}
    \log{\hat{\mathbf{M}}_{i,\cdot}^{h,\tilde{h}}}=
    {\log \mathbf{\hat{J}}}_{i,\cdot}^{h,\tilde{h}}-
    \text{LSE}\left({\log \mathbf{\hat{J}}}_{i,\cdot}^{h,\tilde{h}}\right),
\end{equation}
which is equivalent to the row-normalization of $\hat{\mathbf{J}}^{h,\tilde{h}}$ in the log domain.
The projected logits $\log{\hat{\mathbf{p}}}^{h\rightarrow\tilde{h}}$ can be used to calculate additional \ac{CCE} losses that can contribute to the final loss \cite{spie2023}. However, this does not explicitly model the hierarchical consensus at the different hierarchical levels, requiring an additional step to ensure the joint classification of the classes.

This modeling choice implies that the projection matrices encode dataset cross-level relationships rather than sample-specific transitions. When projecting from a coarse class to finer classes, the projector models global relations in the training distribution, while the direct target-level classifier retains the spectral-temporal characteristics and uncertainty of the individual sample. This formulation provides a compact and interpretable hierarchy-aware regularization mechanism, avoiding the modeling of sample-dependent projectors.

\subsubsection{Semantics-Aware Hierarchical Consensus}
Given the projected prediction at each hierarchical level, we can now define a \ac{HC} class probability vector $\hat{\mathbf{p}}^{\tilde{h}}_{HC}{(\mathbf{x})}$:
\begin{align}
    \hat{\mathbf{p}}^{\tilde{h}}_{HC}{\left(\mathbf{x}\right)} = \notag\\ 
    & \hspace{-5em} = \frac{1}{Z^{\tilde{h}}}\sqrt[H]{\hat{\mathbf{p}}^{\tilde{h}}{\left(\mathbf{x}\right)}
    \prod_{h\neq\tilde{h}} \hat{\mathbf{p}}^{h\rightarrow\tilde{h}}{\left(\mathbf{x}\right)}} \notag\\
    & \hspace{-5em} = \frac{1}{Z^{\tilde{h}}}\sqrt[H]{\hat{\mathbf{p}}^{\tilde{h}}{\left(\mathbf{x}\right)}
    \prod_{h\neq\tilde{h}} (\mathbf{\hat{M}}^{h,\tilde{h}})^{\mathsf{T}} \cdot\hat{\mathbf{p}}^{h}{\left(\mathbf{x}\right)}},
\end{align}
where $\hat{\mathbf{p}}^{\tilde{h}}_{HC}$ is the \ac{HC} class-posterior probability defined as the geometric mean of the predictions at the target hierarchical level $\tilde{h}$ and the projected predictions of all the other $h$ levels, and $Z^{\tilde{h}}$ is a normalization constant to ensure that $\mathbf{p}^{\tilde{h}}_{HC}$ sums to one. This is equivalent to computing the average in the log domain \cite{logpool}:
\begin{align}
    g^{\tilde{h}}_{HC_{j}} &= \frac{1}{H} \left[\log{\hat{P}^{\tilde{h}}{\left(\omega^{\tilde{h}}_{j}|\mathbf{x}\right)}} + 
    \sum_{h \neq \tilde{h}} \log{\hat{P}^{h \rightarrow \tilde{h}}{\left(\omega^{\tilde{h}}_{j}|\mathbf{x}\right)}}\right] \notag \\
    &= \frac{1}{H} \bigg\{
        \left[
            g_{j}^{\tilde{h}} - \text{LSE}\left(\mathbf{g}^{\tilde{h}}\right)
        \right]+ \notag \\ 
        &+ \sum_{h \neq \tilde{h}} \left[
            \text{LSE}\left(\mathbf{g}^{h} + \log{\mathbf{\hat{M}}_{\cdot,j}^{h,\tilde{h}}}\right) - \text{LSE}\left(\mathbf{g}^{h}\right)
        \right]
    \bigg\}, \notag
\end{align}
\begin{equation} \label{eq:hc-computation}
    \hspace{-5em} \log{\hat{\mathbf{p}}^{\tilde{h}}_{HC}{\left(\mathbf{x}\right)}}=
    \mathbf{g}^{\tilde{h}}_{HC}-\text{LSE}{\left(\mathbf{g}^{\tilde{h}}_{HC}\right)},
\end{equation}
where $\text{LSE}{(\mathbf{g}^{\tilde{h}}_{HC})}=\log{Z^{\tilde{h}}}$. This \ac{HC} represents the central \ac{SAHC} consensus found by the voting committee and can serve as an optional hierarchy-aware prediction during inference. The direct prediction $\hat{\mathbf{p}}^{\tilde h}$ preserves sample-specific evidence from the target-level classifier, while the consensus promotes agreement with the remaining levels. The normalized geometric mean is selected because the hierarchical consensus is intended to emphasize agreement among the committee members rather than to provide an average. This differs from an arithmetic mean, where a class can maintain high probability also when it is strongly supported by only one committee member despite disagreement from the others.

\subsubsection{Consensus Loss}
After calculating the consensus, both the prediction at the target level and the projected predictions are compared with the \ac{HC}, defining an additional loss based on the divergence from the \ac{HC}. To this purpose, we employ a \ac{JSD} loss rather than a standard \ac{KLD}. While the \ac{KLD} is widely used to measure the difference between two probability distributions, its asymmetry makes it better suited for scenarios where one distribution is a fixed ``ground truth" or prior. In this case, all hierarchical levels provide dynamically trained ``soft" votes, with no reference distribution. The symmetry in the \ac{JSD} ensures balanced regularization between the projected predictions and the consensus, also providing stabler gradients due to its bounded nature. In the proposed framework, the \ac{JSD} loss is computed as follows:

\begin{multline}
    \text{JSD}^{h,\tilde{h}}\left(\hat{\mathbf{p}}^{\tilde{h}}_{HC}||\hat{\mathbf{p}}^{h\rightarrow\tilde{h}}\right) = \\ 
    = \frac{1}{2} \left(\text{KLD}\left(\hat{\mathbf{p}}^{\tilde{h}}_{HC}||\mathbf{q}\right) + \text{KLD}\left(\hat{\mathbf{p}}^{h\rightarrow\tilde{h}}||\mathbf{q}\right)\right),
\end{multline}
where $\mathbf{q}$ is a mixture distribution of $\hat{\mathbf{p}}^{\tilde{h}}_{HC}$ and $\hat{\mathbf{p}}^{h\rightarrow\tilde{h}}$ calculated as $\mathbf{q}=\frac{1}{2}(\hat{\mathbf{p}}^{\tilde{h}}_{HC}+\hat{\mathbf{p}}^{h\rightarrow\tilde{h}})$, $\text{KLD}$ represents the Kullback-Leibler divergence, and we omit the dependency on $\mathbf{x}$ for simplicity. Please note that when $h=\tilde{h}$, we have that $\hat{\mathbf{p}}^{h\rightarrow\tilde{h}} = \hat{\mathbf{p}}^{\tilde{h}}$. Lastly, we can retrieve the \ac{JSD} losses calculated at the different hierarchical levels and compute the final $\mathcal{L}_{HC}$ penalty loss:

\begin{equation} \label{eq:hc}
    \mathcal{L}_{HC} = \frac{1}{N}\sum_{n=1}^{N}
    \sum_{\tilde{h}=1}^{H}\frac{1}{\log{|\Omega^{\tilde{h}}|}}
    \sum_{h=1}^{H}{\text{JSD}^{h,\tilde{h}}\left(\mathbf{x}_{n}\right)},
\end{equation}
where the logarithm of the number of classes at level $\tilde{h}$, \textit{i.e.}, $|\Omega^{\tilde{h}}|$, allows the balancing of the different number of classes at each hierarchical level. This prevents skewness towards levels with larger number of classes. The $\mathcal{L}_{HC}$ measures the discrepancy between the \ac{HC} prediction and the projected predictions, encouraging the different layers to converge to the committee's vote.

Finally, we also apply full supervision on the consensus predictions at the different levels considering a standard \ac{CCE} loss:
\begin{equation} \label{eq:hccce}
    \mathcal{L}_{CCE}^{HC} = \sum_{h=1}^{H}-\frac{1}{N}\sum_{n=1}^{N}\sum_{i=1}^{|\Omega^{h}|} \mathbbm{1}\left[y^{h}_{n}=\omega^{h}_{i}\right]\log\hat{P}^{h}_{HC}\left(\omega^{h}_{i}|\mathbf{x}_{n}\right).
\end{equation}

\subsection{Overall Loss Function}
Let $\theta_\mathbf{f}$ be the parameters of the backbone feature extractor $(\mathbf{f};\theta_\mathbf{f})$, and let $\theta_{\mathbf{g}^h}$ be the parameters of the classification head $({\mathbf{g}^h};\theta_{\mathbf{g}^h})$ associated with level $h$. Moreover, let $\mathcal W=\{\mathbf W^{a,b}\}_{1\leq a<b\leq H}$ be the set of trainable hierarchy-parameter matrices. The complete set of trainable parameters can be defined as follows.
\begin{equation}
    \Theta=\left\{\theta_\mathbf{f}\right\}\cup\{\theta_\mathbf{g}^h\}_{h=1}^{H}\cup\mathcal W.
\end{equation}
The final optimization problem is defined as:
\begin{equation}
    \Theta^\star = \arg\min_{\Theta} \mathcal L(\Theta),
\end{equation}
where the final loss function $\mathcal{L}$ is derived from the sum of the losses computed in \eqref{eq:cce}, \eqref{eq:hc}, and \eqref{eq:hccce}:
\begin{multline}
    \mathcal{L}\left( \Theta \right) = \mathcal{L}_{CCE}\left( \theta_\mathbf{f}, \{\theta_\mathbf{g}^h\}_{h=1}^{H} \right) + \\ + \lambda_{HC}(\mathcal{L}_{CCE}^{HC}\left( \theta_\mathbf{f},\{\theta_\mathbf{g}^h\}_{h=1}^{H}, \mathcal W \right) + \\ + \mathcal{L}_{HC}\left(\theta_\mathbf{f}, \{\theta_\mathbf{g}^h\}_{h=1}^{H}, \mathcal W \right)),
\end{multline}
where $\lambda_{HC}$ is a metaparameters tuning the contribution of the \ac{HC} losses. It is important to note that $\mathcal{L}_{HC}$ does not require direct supervision, as it is simply a loss representing the divergence from the \ac{HC}. The \ac{HC} prediction is used both within the penalty loss and the additional supervised \ac{CCE} loss during the training phase. The \ac{HC} prediction at each hierarchical level can also be used at inference to exploit the cross-level agreement learnt during self-consistent training, supporting coherent hierarchical predictions. The pseudocode of the proposed framework is detailed in Alg. \ref{alg:sahc}.

\begin{algorithm}[t]
\color{black}
\caption{\textcolor{black}{Proposed SAHC training approach}}
\label{alg:sahc}
\begin{algorithmic}[1]
\REQUIRE Training set \(\mathcal D\), hierarchical levels
\(\{\Omega^h\}_{h=1}^{H}\), indicator matrices
\(\{\mathbf I^{a,b}\}_{1\leq a<b\leq H}\), loss weights
\(\{\lambda^h\}_{h=1}^{H}\), \(\lambda_{\mathrm{HC}}\), warm-up schedule
\(\alpha(t)\).
\ENSURE Optimized parameters
\(\Theta=\left\{\theta_\mathbf{f}\right\}\cup\{\theta_\mathbf{g}^h\}_{h=1}^{H}\cup\mathcal W\).

\STATE Initialize the backbone parameters \(\theta_\mathbf{f}\) and the
classification-head parameters \(\{\theta_\mathbf{g}^h\}_{h=1}^{H}\).
\STATE Initialize each hierarchy-parameter matrix
\(\mathbf W^{a,b}\in\mathcal W\) from the corresponding fixed indicator
matrix \(\mathbf I^{a,b}\) according to \eqref{eq:wfromindicator}.
\FOR{each training epoch \(t\)}
    \FOR{each mini-batch \(\mathcal B\subset\mathcal D\)}
        \STATE Compute features
        $
        \mathbf z=\mathbf{f}(\mathbf x;\theta_\mathbf{f}), \quad \mathbf x\in\mathcal B .
        $
        \STATE For each level \(h\), compute logits $\mathbf r^h$ and probabilities $\hat{\mathbf p}^{h}$:
        \[
        \mathbf r^h=\mathbf{g}^h(\mathbf z;\theta_\mathbf{g}^h), \qquad
        \hat{\mathbf p}^{h}=\sigma(\mathbf r^h).
        \]
        \STATE Compute the multi-level supervised loss
        \(\mathcal L_{\mathrm{CCE}}\) using \eqref{eq:cce}.
        \STATE For each hierarchy pair \((a,b)\), compute
         \(\log\mathbf J^{a,b}\) and the corresponding
        conditional projection matrices
        \(\log\mathbf M^{h,\tilde h}\).
        \FOR{each target level \(\tilde h=1,\ldots,H\)}
            \STATE Project the predictions from all source levels
            \(h\neq \tilde h\) to level \(\tilde h\) using \eqref{eq:proj-log}.
            \STATE Compute the hierarchical consensus prediction
            \(\hat{\mathbf p}_{\mathrm{HC}}^{\tilde h}\) using \eqref{eq:hc-computation}.
        \ENDFOR
        \STATE Compute the consensus loss
        \(\mathcal L_{\mathrm{HC}}\) using \eqref{eq:hc}.
        \STATE Compute the supervised consensus loss
        \(\mathcal L_{\mathrm{CCE}}^{\mathrm{HC}}\) using \eqref{eq:hccce}.
        \STATE Minimize
        \[
        \mathcal L =
        \mathcal L_{\mathrm{CCE}}
        +
        \alpha(t)\lambda_{\mathrm{HC}}(\mathcal L_{\mathrm{CCE}}^{\mathrm{HC}}
        +
        \mathcal L_{\mathrm{HC}}) .
        \]
        \STATE Update \(\theta_\mathbf{f}\), \(\{\theta_\mathbf{g}^h\}_{h=1}^{H}\), and
        \(\mathcal W\) by back-propagation.
    \ENDFOR
\ENDFOR
\end{algorithmic}
\end{algorithm}

\subsection{Computational Complexity and Scalability}
The \ac{SAHC} formulation learns a hierarchy parameter matrix for every unordered pair of hierarchical levels. Let $|\Omega^h|$ denote the number of classes at level $h$. Although the number of hierarchy matrices is $H(H-1)/2$, their practical memory requirement depends on the cardinality of the corresponding label spaces. The total number of trainable hierarchy parameters is:
\begin{equation}
    P_{\mathrm{hier}}=\sum_{1\leq h_1<h_2\leq H} |\Omega^{h_1}||\Omega^{h_2}|.
\end{equation}
The matrices $\log\mathbf{J}^{h_2,h_1}$ and $\log\mathbf{M}^{h,\tilde h}$ are differentiable quantities derived from the trainable matrices $\mathbf{W}^{h_2,h_1}$. Thus, they do not introduce additional independent parameters. The additional parameters associated with the level-specific linear heads scale as $\mathcal{O}(d\sum_{h=1}^{H}|\Omega^{h}|)$, where $d$ is the backbone feature dimension. For a mini-batch of size $B$, the consensus requires a projection from each source level to each distinct target level. Since the projection in \eqref{eq:proj-nolog} has complexity of $\mathcal{O}(B|\Omega^{h}||\Omega^{\tilde h}|)$, the overall cross-level projection cost is:
\begin{equation}
    \mathcal{O}\left(2BP_{\mathrm{hier}}\right).
\end{equation}
The consensus fusion and \ac{JSD} operate on probability vectors and scale as $\mathcal{O}(BH\sum_{h=1}^{H}|\Omega^{h}|)$. Therefore, when all levels have an average cardinality $\bar {|\Omega|} = C$, the implementation has a quadratic dependence on hierarchy depth, \textit{i.e.}, $\mathcal{O}(H^2C^2)$ for the hierarchy parameters and cross-level projections. For a deep \ac{RS} hierarchy, as the one analyzed later in this study before class filtering, a hierarchy with $H=4$ and $(|\Omega^{1}|,|\Omega^{2}|,|\Omega^{3}|,|\Omega^{4}|)=(5,15,40,90)$ classes per-level, the complete formulation contains six trainable matrices and $6275$ parameters, usually negligible relative to the backbone parameters and the added projection cost remains modest.

\section{Dataset Description and Experiment Setup}\label{dataset}

This Section provides a description of the datasets employed in this study. Then, the experimental setup, including the preprocessing phase and the benchmark methodologies adopted for evaluating the \ac{DL} models, are described.

\subsection{NWPU-RESISC45}
To assess the performance of the proposed methodology on \ac{VHR} images, where spatial features play a prominent role, we considered the NWPU-RESISC45 \cite{cheng2017remote} benchmark as a first dataset. NWPU-RESISC45 is composed of $31.500$ RGB multi-resolution images for scene classification of \ac{VHR} images categorized into 45 different classes at the finest-level of detail, and exhibits a high within-class diversity and between-class similarity. The dataset is completely balanced, \textit{i.e.}, each class has the same number of samples at the \textit{fine-grained} level. We defined a custom-hierarchy based on the subdivision defined in MillionAID \cite{long2021creating}, resulting in 8, 24, and 45 classes, corresponding to \textit{coarse-grained}, \textit{intermediate}, and \textit{fine-grained} classes, respectively. 

\subsection{ELU Dataset}\label{UDS_emilia}
\begin{figure*}
    \centering
    \includegraphics[width=\textwidth]{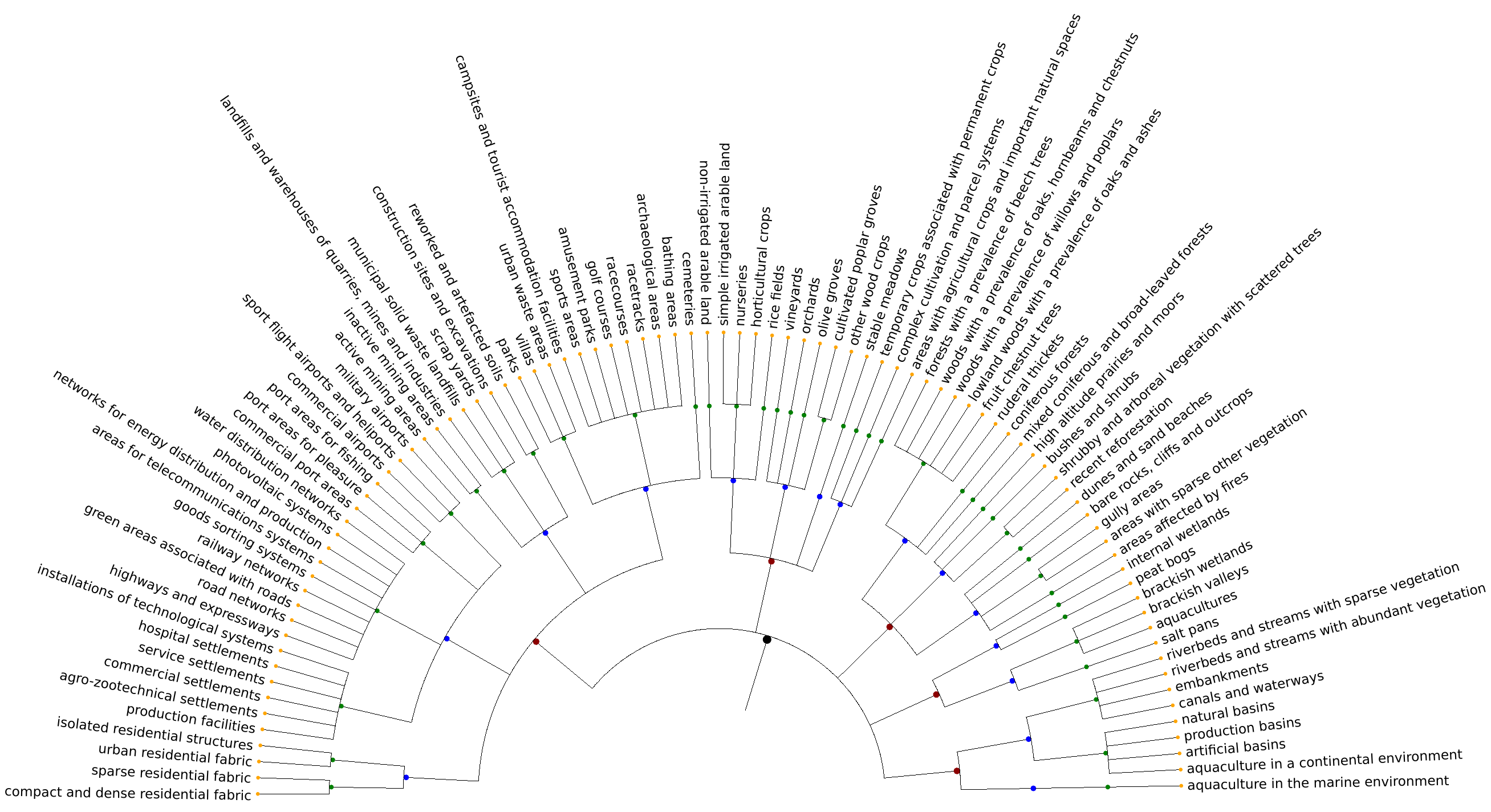}
    \caption{A semicircular hierarchical label tree representation of the \ac{ELU} dataset at the fine-grained level of the hierarchy. The leaves corresponding to the fine-grained classes are shown in orange, while the nodes for intermediate classes are depicted in green, those for coarse-grained classes are shown in blue, and the nodes for macro in brown. The root node is represented in black.}
    \label{fig:tree}
\end{figure*}
To further validate the performance of the proposed methodology and demonstrate its generalizability on different land cover mapping tasks, we additionally considered a multitemporal multispectral dataset differing in terms of complexity in the hierarchical structure. This dataset\footnote{[Online]. Available: \footlink{https://geoportale.regione.emilia-romagna.it/catalogo/dati-cartografici/pianificazione-e-catasto/uso-del-suolo/layer-14}{}. Accessed April 9, 2026.} consists of a publicly available \ac{LU} map of $2020$ over the Emilia Romagna region, Italy. The maps were defined considering the standard \ac{CLC} aggregation at four levels, showcasing a highly detailed semantic dataset. For the purpose of this analysis, we evaluated the hierarchical approach on four levels of hierarchy, consisting of the first three levels of the \ac{CLC} and the additional finest level of detail defined by the \ac{LULC}, here referred to as \textit{macro}, \textit{coarse-grained}, \textit{intermediate} and \textit{fine-grained}, respectively. The dataset has been built considering orthophotos and infrared images, and shows a minimum mapping unit of $0.16$ hectares, with a minimum of $7\text{m}$ for linear structures.
Additionally, it represents \ac{LU} classes, where the same \ac{LC} class could be associated with different semantic classes, such as the fine-grained classes ``Commercial airports'', ``Sport flight airports and heliports'', and ``Military airports''. We considered the \ac{ELU} dataset corresponding to the \ac{MGRS} \ac{S2} granule \textit{32TPQ} ($10980\times10980$ pixels, for a total area of $110\times110$ $km^{2}$). In the analyzed  granule, the four hierarchical levels show $5$, $15$, $40$, and $90$ classes at the macro, coarse-grained, intermediate, and fine-grained levels, respectively. Please note that such an elevated number of fine-grained classes is not commonly approached in \ac{EO} classification problems, where the majority of studies focus on the intermediate or coarse-grained classification task. Nonetheless, this level of precision allows a better monitoring of the class of interest and the extraction of information-rich content. Due to the high number of classes and, consequently, the large size of the hierarchical tree, we refer to this hierarchy as \textit{a complex} hierarchy. A semicircular tree representing the hierarchical structure of the dataset is shown in Fig. \ref{fig:tree}.

\subsection{S2 Preprocessing and Dataset Preparation}
\subsubsection{VHR dataset} 
The training, validation, and test splits were selected accordingly to \cite{9324501}, considering a split-ratio of $0.6$, $0.2$, and $0.2$, respectively. During training, we applied data augmentation techniques consisting of resize, random crops and random rotations of the images.

\subsubsection{Multitemporal MS datasets}
The input data consists of \ac{S2} \ac{MS} satellite images provided in the Copernicus \ac{EO} program. The \ac{S2} acquisitions considered have been preprocessed at \ac{L2A} and re-projected at $10\text{m}$/pixel resolution through a nearest neighbor approach. In the experiments, all the bands except the $60\text{m}$ resolution band were used. Due to the high-complexity of this classification problem, we collected a \ac{TS} of \ac{S2} images, fully exploiting the temporal signature and allowing the detection of seasonal targets. The \ac{S2} \ac{TS} considered for the Emilia Dataset consists of acquisitions ranging from January $1^\text{st}$ $2020$ to December $31^\text{st}$ $2020$. The \ac{S2} \ac{TS} has been processed to achieve a more regular \ac{TS}, resulting in $12$ medoid monthly composites \cite{S2GM_Service}. 

To extract a balanced training set, we considered an approach based on the heterogeneity of the input training data. In particular, we divided the initial \ac{S2} monthly composite images into non-overlapping tiles of $1000\times1000$ pixels ($100km^{2}$). After the initial subdivision, we randomly subdivided the tiles into training, validation, and test sets by considering a split-ratio of $0.7$, $0.15$ and $0.15$, respectively. We extracted $N=2000$ representative points for each tile and class in the analyzed classification scheme. Classes not present in all sets were excluded from the analysis, resulting in $74$, $33$, $13$, and $5$ classes for the different hierarchical levels. In the case of minority classes having less than $N$ samples, all the available samples were extracted. The training set was generated by extracting patches of $9\times9$ pixels around the selected points in the training tiles. To minimize spatial autocorrelation between the training patches, the corresponding reference data was extracted only for the central $3\times3$ pixels. In this way, we collected a training dataset representative of the entirety of the classification scheme, corresponding to $\sim1.17\%$ ($1.413.522$ pixels) of the total number of pixels available in the \ac{MGRS} \ac{S2} granule. 

\subsection{Design of Experiments}
Validation has been conducted on both datasets considering different multi-granularity approaches. The reported results include the values of \ac{OA} and \ac{mF1} for both tasks, while for the multispectral dataset also Top-3 and Top-5 metrics are reported. We considered different multi-granularity modules as comparison, all built on the same backbone architecture for fairness on all the datasets. We considered the following approaches: 
\begin{enumerate}
    \item \textbf{Baseline}: the baseline trained only with the level-specific hierarchical head through standard \ac{CCE} minimization. Selected to evaluate the improvement gained by incorporating the hierarchy.
    \item \textbf{HAF} \cite{garg2022learning}: selected as a regularization approach adopting soft labels and geometric constraints that use auxiliary classifiers at each level to ensure consistent predictions with the hierarchy, specifically aiming to reduce the severity of mistakes.
    \item \textbf{HRN} \cite{chen2022label}: included as state-of-the-art benchmark that facilitates hierarchical feature interaction by forcing finer classes to inherit attributes from coarser classes through residual skip connections.
    \item \textbf{MGM} \cite{Zhou_Wei_Zhang_Qi_Yang_Li_2023}: included to assess the effectiveness of using knowledge-guided relation graphs and bi-directional interaction structures that enhance coarse-grained features through gradient truncations. 
    \item \textbf{HiCervix} \cite{hicervix}: chosen as benchmark for methods that model semantic correlations within an explicit tree structure using soft labels and recursive probability summation across hierarchical paths.
    \item \textbf{ALGS} \cite{chen2025adaptive}: selected to evaluate the impact of a global semantic distance-based penalty term for adjusting class predictions, ensuring hierarchical consistency when classification is enforced at every level. Specifically designed for \ac{RS} tasks.
    \item \textbf{HCL} \cite{chen2025hierarchical}: selected as comparison for representation-based learning that disentangles class-specific embeddings at multiple levels to align them with their semantic relationships. Specifically designed for \ac{RS} tasks.
\end{enumerate}

For the \ac{VHR} task, we considered a ResNet50 backbone pretrained with ImageNet weights \cite{pytorchresnet}. This choice is due to its deeply stacked convolutional layers and residual connections, making it well-suited for extracting highly localized, fine-grained spatial and morphological features necessary for \ac{VHR} analysis. For the \ac{MS} \ac{SITS} classification task, we adopted a different backbone architecture designed to process multispectral and temporal information. This backbone consists of an adaptation to the \ac{RS} scenario of a \textit{Swin Transformer} \cite{geng2022rstt} and is trained from scratch. For a detailed description of the layer topology, please refer to the cited paper \cite{geng2022rstt}. The choice of this backbone is due to its shifted-window self-attention mechanism that efficiently captures both local details and long-range global contextual dependencies, making it ideal for processing complex multi-scale structural variations.

We adopted the \textit{AdamW} optimizer for all the architectures and configurations. The metaparameters of each different configuration have been validated considering a \ac{TPE} algorithm to efficiently explore the metaparameters space. The learning rate and the weight decay were sampled from log-uniform distributions $U_{\log}([10^{-6},10^{-3}])$. We considered 50 training epochs on both datasets, together with \ac{SWA} to enhance model generalization \cite{izmailov2019averagingweightsleadswider}. Considering the proposed \ac{SAHC} approach, the weight $\lambda_{HC}$ associated to the \ac{HC} loss has been sampled from a log-uniform distribution $U_{\log}([10^{-2},10^{2}])$, while the weights $\lambda^{h}$ have been set to $0.5$, $0.2$, and $0.3$ for the \ac{VHR} dataset, and to $0.4$, $0.2$, $0.1$, and $0.3$ for the \ac{MS} dataset, to prioritize the fine-grained classification and reduce the influence of the intermediate, coarse-grained, and macro levels. For the NWPU-RESISC45 dataset, these weights were selected on the validation set through a simplex grid search. For the ELU dataset, we heuristically fixed the weights according to the trend observed in the NWPU-RESISC45. In both datasets, the learnable parameter matrices $\mathbf{W}^{h,\tilde{h}}$ were initialized with $\Delta=5$ and $s=0.01$. We implemented a dynamic consistency warmup schedule for the \ac{HC} loss weight, $\lambda_{HC}$. Specifically, during the training phase, $\lambda_{HC}$ was initially fixed to 0 for the first 5 epochs to allow for stable feature extraction. Subsequently, the weight value was linearly ramped up from epoch 5 to 15 until reaching its sampled maximum value, where it remained constant for the duration of the training. We report the average scores on three runs, together with their unbiased estimation of standard deviation.

\section{Experimental Results}\label{results}
\begin{table*}
\centering
\caption{Comparison of the Accuracies obtained by different hierarchical methodologies on the NWPU-RESISC45 dataset. The best results are reported in bold, second best are underlined. The related standard deviation is reported in parentheses}
\label{tab:nwpuresisc}
\begin{tblr}{
  colspec = {l l *{8}{X[c]}},
  row{2-11} = {fg=black},
  row{1} = {m},
  cell{2}{1} = {r=3}{},
  cell{2}{10} = {Alto,font=\bfseries},
  cell{3}{10} = {Alto,font=\bfseries},
  cell{4}{10} = {Alto,font=\bfseries},
  cell{5}{1} = {r=3}{},
  cell{5}{10} = {Alto,font=\bfseries},
  cell{6}{10} = {Alto,font=\bfseries},
  cell{7}{10} = {Alto,font=\bfseries},
  vline{3} = {1-11}{},
  hline{1-2,8,12} = {-}{},
  hline{5} = {dashed},
  hline{1,12} = {2}{-}{},
  cell{12}{1} = {c=10}{},
}
Metric & Hierarchy & Baseline \cite{pytorchresnet} & HRN \cite{chen2022label} & HAF \cite{garg2022learning} & ALGS \cite{chen2025adaptive} & HCL \cite{chen2025hierarchical} & MGM \cite{Zhou_Wei_Zhang_Qi_Yang_Li_2023} & HiCervix \cite{hicervix} & SAHC\\
OA (\%) & Fine-grained & 96.12~(0.12) & \uline{96.34~(0.19)} & 96.05~(0.13) & 96.28~(0.06) & 95.70~(0.16) & 96.20~(0.10) & 95.55~(0.48) & 96.55~(0.08)\\
 & Intermediate & 97.29~(0.11) & 97.30~(0.14) & 97.13~(0.18) & \uline{97.41~(0.05)} & 97.00~(0.22) & 97.27~(0.23) & 96.88~(0.27) & 97.48~(0.07)\\
 & Coarse-grained & 97.43~(0.08) & 97.82~(0.11) & 97.76~(0.09) & \uline{97.85~(0.11)} & 97.57~(0.16) & 97.79~(0.25) & 97.50~(0.20) & 97.90~(0.07)\\
mF1(\%) & Fine-grained & 96.11~(0.11) & \uline{96.34~(0.19)} & 96.05~(0.13) & 96.28~(0.06) & 95.70~(0.15) & 96.20~(0.11) & 95.55~(0.48) & 96.56~(0.07)\\
 & Intermediate & 96.59~(0.10) & 96.56~(0.19) & 96.20~(0.23) & \uline{96.65~(0.04)} & 96.05~(0.19) & 96.52~(0.42) & 95.97~(0.35) & 96.75~(0.16)\\
 & Coarse-grained & 97.07~(0.10)  & 97.50~(0.17) & 97.39~(0.11) & \uline{97.55~(0.10)} & 97.16~(0.18) & 97.41~(0.31) & 97.10~(0.26) & 97.58~(0.09)\\
Params (M) &  & 25.8 & 35.5 & 25.9 & 25.9 & 27.0 & 35.5 & 25.9 & 25.9\\
FLOP (T) &  & 6.22 & 7.03 & 6.22 & 6.22 & 6.51 & 6.88 & 6.22 & 6.22\\
Inference time\textsuperscript{a} (ms) &  & 135.48 & 139.23 & 135.68 & 135.62 & 138.21 & 139.30 & 135.73 & 135.62\\
Peak VRAM\textsuperscript{a} (MB) &  & 2811.16 & 2851.75 & 2812.60 & 2812.42 & 2817.01 & 2851.88 & 2812.91 & 2812.24\\
\textsuperscript{a} Results for a single forward pass over a batch of 256 images on an NVIDIA L40S GPU. Inference time is averaged over 50 forward passes.
\end{tblr}
\end{table*}
The experiments focus on validating the proposed \ac{SAHC} approach (with \ac{HC} predictions at inference time) over different datasets, comparing it to other state-of-the-art hierarchical approaches on different levels of the hierarchy. We first focus on the results on the specific tasks, \textit{i.e.}, scene classification on the \ac{VHR} dataset and semantic segmentation of \ac{MS} \ac{SITS}. After this, we focus on ablation studies on the divergence module and the fusion operator. Then, we analyze the importance of the hierarchical weights selection $\lambda^{h}$ on the NWPU dataset. Subsequently, a complete ablation study on both datasets on each component of the proposed \ac{SAHC} is conducted, as well as a sensitivity analysis on the initializations of the projectors. Finally, we discuss a visualization of the trained hierarchy projection matrices to identify the key hierarchical relationships learnt by the proposed approach.

\subsection{Case I: VHR Scene Classification}\label{subsec:caseelu}
The performance of the proposed \ac{SAHC} method on the NWPU-RESISC45 dataset is summarized in Table \ref{tab:nwpuresisc}, providing a comparative analysis against state-of-the-art methodologies and the non-hierarchical baseline. Given the balanced class distribution characteristic of the NWPU-RESISC45 dataset, the \ac{OA} and \ac{mF1} are aligned across all evaluated models. From the table, one can see that the integration of hierarchical information consistently improves classification performance over the baseline at each hierarchical level. Indeed, the evaluated hierarchical architectures outperform their non-hierarchical counterpart, confirming that the exploitation of semantic relationships allows for a more robust distinction between fine-grained \ac{RS} classes. Among the compared methodologies, the proposed \ac{SAHC} achieves the highest performances at the fine-grained level, with an \ac{OA} of $96.55$\% and an \ac{mF1} of $96.56$\%. Notably, the good performance of \ac{SAHC} extends across the entire taxonomic hierarchy. At the intermediate and coarse-grained levels, the proposed method achieves \acp{OA} of $97.48$\% and $97.90$\%, respectively, consistently outperforming the other frameworks. The second-best overall performance is obtained by the \ac{HRN} method at the most detailed hierarchical level, which reaches a fine-grained \ac{OA} of $96.34$\%. At coarsest levels, \ac{ALGS} outperforms the other state-of-the-art methodologies, while still performing slightly lower compared to the proposed \ac{SAHC}. This comparison also highlights the computational overhead introduced by the different architectures. The \ac{HRN} method achieves its high accuracy at the cost of architectural efficiency. Indeed, the introduction of the \ac{GSB} leads to a significant increase in parameters and computational overload. In contrast, the proposed \ac{SAHC} maintains nearly identical computational footprint as the baseline architecture. This efficiency is shared by other methods such as \ac{ALGS}, \ac{HAF}, and HiCervix, where the bottleneck remains the backbone itself. In the specific case of \ac{SAHC}, the architectural overhead is due to the three auxiliary heads and their associated projection matrices, both of which are negligible relative to the dimension of the backbone. When focusing strictly on architectures that do not introduce additional complexity, the \ac{ALGS} method provides the highest performance after \ac{SAHC}, though it remains lower in terms of \ac{OA}. 

It is worth noting that NWPU-RESISC45 represents a highly saturated benchmark in this experimental setting, as the non-hierarchical baseline already achieves $96.12$\% \ac{OA} and $96.11$\% \ac{mF1} at the fine-grained level. Consequently, large absolute gains are not expected. Nevertheless, SAHC achieves the best mean performance at all hierarchical levels in both \ac{OA} and \ac{mF1}, indicating that the hierarchical consensus improves the fine-grained prediction without degrading the intermediate and coarse-grained outputs. At the fine-grained level, SAHC improves the \ac{OA} from $96.12$\% to $96.55$\% and the \ac{mF1} from $96.11$\% to $96.56$\%. Moreover, the standard deviation of SAHC at the fine-grained level is $0.08$\% for \ac{OA} and $0.07$\% for \ac{mF1}, showing that the improvement is stable across runs. These results suggest that, even when the flat baseline is already very strong, the proposed hierarchical consensus provides a consistent improvement while preserving computational efficiency.

\subsection{Case II: MS SITS Semantic Segmentation}

\begin{table*}
\centering
\caption{Comparison of the Accuracies obtained by different hierarchical methodologies on the ELU dataset. The best results are reported in bold, second best are underlined. The related standard deviation is reported in parentheses}
\label{tab:semanticsegm}
\begin{tblr}{
  colspec = {l l *{8}{X[c]}},
  row{2-22} = {fg=black},
  row{1} = {m},
  cell{2}{1} = {r=4}{},
  cell{2}{10} = {Alto,font=\bfseries},
  cell{3}{10} = {Alto,font=\bfseries},
  cell{4}{10} = {Alto,font=\bfseries},
  cell{5}{5} = {Alto,font=\bfseries},
  cell{6}{1} = {r=4}{},
  cell{6}{10} = {Alto,font=\bfseries},
  cell{7}{10} = {Alto,font=\bfseries},
  cell{8}{10} = {Alto,font=\bfseries},
  cell{9}{9} = {Alto,font=\bfseries},
  cell{10}{1} = {r=4}{},
  cell{10}{10} = {Alto,font=\bfseries},
  cell{11}{10} = {Alto,font=\bfseries},
  cell{12}{10} = {Alto,font=\bfseries},
  cell{14}{1} = {r=4}{},
  cell{14}{10} = {Alto,font=\bfseries},
  cell{15}{10} = {Alto,font=\bfseries},
  cell{16}{10} = {Alto,font=\bfseries},
  cell{17}{5} = {Alto,font=\bfseries},
  vline{3} = {1-21}{},
  hline{1-2,18,22} = {-}{},
  hline{6,10,14} = {dashed},
  hline{1,22} = {2}{-}{},
  cell{22}{1} = {c=10}{},
}
Metric & Hierarchy & Baseline \cite{pytorchresnet} & HRN \cite{chen2022label} & HAF \cite{garg2022learning} & ALGS \cite{chen2025adaptive} & HCL \cite{chen2025hierarchical} & MGM \cite{Zhou_Wei_Zhang_Qi_Yang_Li_2023} & HiCervix \cite{hicervix} & SAHC\\
OA (\%) & Fine-grained & 36.84~(0.37) & 36.49~(0.19) & 36.11~(0.69) & 35.94~(0.07) & \uline{37.58~(0.30)} & 35.54~(0.27) & 37.30~(0.83) & 38.38~(0.31)\\
 & Intermediate & 47.84~(0.46) & 46.57~(0.03) & 39.72~(1.73) & 46.18~(0.23) & \uline{49.20~(0.37)} & 46.45~(0.43) & 48.18~(0.65) & 50.20~(0.13)\\
 & Coarse-grained & 57.64~(0.07) & 56.79~(0.30) & 57.94~(0.50) & 56.29~(0.12) & \uline{59.86~(0.25)} & 56.53~(0.28) & 58.55~(0.51) & 60.21~(0.21)\\
 & Macro & 77.55~(0.04) & 75.95~(0.44) & 78.88~(0.14) & 75.63~(0.12) & \uline{78.47~(0.19)} & 75.76~(0.39) & 77.52~(0.11) & 78.33~(0.17)\\
Top-3 (\%) & Fine-grained & 56.78~(0.03) & 56.50~(0.94) & 54.84~(1.37) & 54.83~(0.08) & 57.29~(0.31) & 56.12~(0.27) & \uline{57.97~(1.42)} & 59.33~(0.35)\\
 & Intermediate & 70.86~(0.50) & 70.27~(0.64) & 65.29~(1.48) & 69.28~(0.32) & \uline{72.00~(0.36)} & 70.17~(0.24) & 71.82~(0.80) & 73.71~(0.23)\\
 & Coarse-grained & 81.35~(0.11) & 81.75~(0.67) & 81.12~(0.13) & 80.79~(0.21) & \uline{83.00~(0.27)} & 81.47~(0.32) & 82.71~(0.68) & 84.49~(0.12)\\
 & Macro & 96.58~(0.08) & 96.89~(0.11) & 95.97~(0.51) & 97.04~(0.15) & \uline{97.21~(0.15)} & 96.79~(0.20) & 97.29~(0.19) & 97.15~(0.17)\\
Top-5 (\%) & Fine-grained & 65.23~(0.13) & 65.25~(1.18) & 63.66~(1.07) & 63.15~(0.03) & 65.82~(0.31) & 65.12~(0.22) & \uline{66.69~(1.65)} & 68.00~(0.32)\\
 & Intermediate & 80.01~(0.27) & 80.06~(0.55) & 75.49~(0.83) & 78.76~(0.40) & 81.05~(0.40) & 79.76~(0.20) & \uline{81.08~(0.80)} & 82.66~(0.14)\\
 & Coarse-grained & 89.45~(0.15) & 90.62~(0.24) & 88.44~(0.55) & 89.93~(0.22) & 90.89~(0.38) & 89.93~(0.29) & \uline{91.08~(0.44)} & 91.88~(0.10)\\
 & Macro & — & — & — & — & — & — & — & —\\
mF1 (\%) & Fine-grained & 32.57~(0.23) & 32.36~(0.44) & 31.40~(1.12) & 31.73~(0.17) & \uline{33.48~(0.22)} & 31.63~(0.19) & 33.12~(0.93) & 34.56~(0.32)\\
 & Intermediate & 40.64~(0.38) & 39.33~(0.09) & 26.34~(1.66) & 38.75~(0.45) & \uline{41.23~(0.33)} & 39.27~(0.18) & 40.32~(1.05) & 42.97~(0.37)\\
 & Coarse-grained & 52.89~(0.14) & 52.11~(0.48) & 48.02~(1.70) & 51.42~(0.02) & \uline{54.94~(0.30)} & 51.79~(0.36) & 53.57~(0.87) & 55.54~(0.15)\\
 & Macro & 68.35~(0.18) & 66.91~(0.43) & 70.56~(0.20) & 66.85~(0.40) & \uline{69.35~(0.58)} & 66.65~(0.54) & 68.54~(0.50) & 69.06~(0.16)\\
Params (M) &  & 4.7 & 16.2 & 4.8 & 4.8 & 5.8 & 16.2 & 4.7 & 4.8\\
FLOP (T) &  & 2.22 & 2.38 & 2.22 & 2.22 & 2.41 & 2.32 & 2.22 & 2.22\\
Inference time\textsuperscript{a} (ms) &  & 81.45 & 120.43 & 119.62 & 119.64 & 121.46 & 120.39 & 119.64 & 84.18\\
Peak VRAM\textsuperscript{a} (MB) &  & 1946.24 & 3404.32 & 3362.73 & 3362.14 & 3308.48 & 3349.73 & 3304.87 & 1958.99\\
\textsuperscript{a} Results for a single forward pass over a batch of 256 patches on an NVIDIA L40S GPU. Inference time is averaged over 50 forward passes.
\end{tblr}
\end{table*}
To further evaluate the robustness of the proposed framework, an extensive analysis was conducted on the highly complex \ac{ELU} dataset. Unlike the previous scenario, this dataset consists of lower-resolution multispectral images ($10m$/pixel) mapped to a complex taxonomy of $74$ classes distributed across four hierarchical levels. The results of this analysis are reported in Table \ref{tab:semanticsegm}. The complexity of the \ac{ELU} dataset, combined with the potential for semantic noise or inconsistencies in the \ac{CLC} \ac{LU}, makes this scenario particularly challenging. Standard hierarchical models that incorporate a rigid taxonomy might struggle to adapt to these inconsistencies. 
The proposed \ac{SAHC} mitigates this limitation through consensus training and taxonomy-guided projectors, that are adaptively tuned during optimization.
This advantage can be observed in the \ac{OA} and \ac{mF1} shown in the Table. Indeed, \ac{SAHC} achieves the best performance at the fine-grained level with an \ac{OA} of $38.38$\% and an \ac{mF1} of $34.56$\%. These results are also confirmed at the intermediate (\ac{OA} $50.20$\%, \ac{mF1} $42.97$\%) and coarse-grained levels (\ac{OA} $60.21$\%, \ac{mF1} $55.54$\%), consistently better performing with respect to other state-of-the-art methodologies like HiCervix and \ac{HCL}. At the macro level, \ac{SAHC} demonstrates competitive performance, ranking after only to the \ac{HCL} and \ac{HAF} methods by a marginal gap.

Although \ac{SAHC} achieves the best fine-grained performance among the compared methods, the absolute \ac{OA} at the finest level remains relatively low. This behavior is mainly due to the intrinsic complexity of the \ac{ELU} dataset. Indeed, \ac{ELU} contains $74$ classes, making the Top-1 classification problem substantially more difficult than standard \ac{EO} classification benchmarks. Then, the taxonomy represents \ac{LU} classes, not only \ac{LC} categories. Therefore, several fine-grained classes may be semantically distinct in the reference map while sharing similar spectral-temporal signatures at \ac{S2} resolution.

\begin{figure*}
  \centering
  \includegraphics[width=\textwidth]{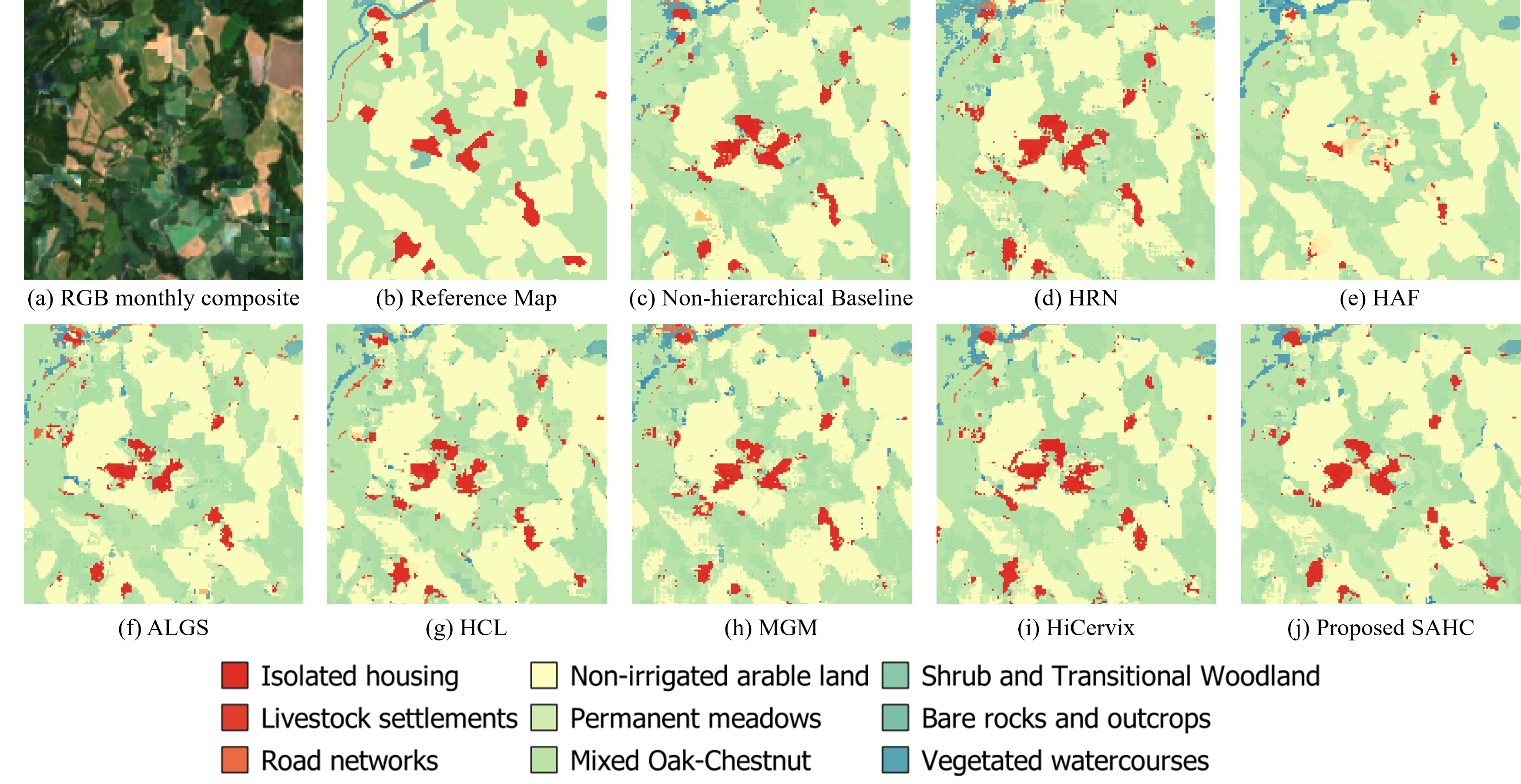}
  \caption{Qualitative example of the predictions at the fine-grained level on the \ac{ELU} dataset obtained by the considered hierarchical methodologies using the Swin Transformer backbone: (a) true color (RGB) monthly composite for August 2020, (b) \ac{CLC} reference map, (c) non-hierarchical baseline (d) HRN method, (e) HAF method, (f) ALGS method, (g) HCL method, (h) MGM method, (i) HiCervix method, and (j) the proposed \ac{SAHC} method.}
  \label{fig:elu_pred}
\end{figure*}

The Top-3 and Top-5 accuracy metrics evaluated at all the levels of the hierarchy further support this interpretation. These metrics indicate whether the ground-truth class is among the highest-confidence predictions of the model, measuring the capacity of the network to maintain semantic proximity even when the absolute Top-1 prediction is incorrect. The proposed \ac{SAHC} approach demonstrates the highest performance across almost all granularity levels. Specifically, \ac{SAHC} achieves the highest Top-3 accuracies for the fine-grained, intermediate, and coarse-grained levels, reaching a $59.33$\%, $73.71$\%, and $84.49$\%, respectively. This represents a large improvement against the \ac{HRN}, which ranked second in the \ac{VHR} case, showing a margin of $2.83$\% at fine-grained level, while also outperforming the \ac{RS}-oriented competitor \ac{HCL} by $2.04$\%, $1.71$\% and $1.49$\% in the fine, intermediate and coarse levels, respectively. At the macro level, Top-3 performance saturates for all evaluated models. In this aggregated space, \ac{SAHC} remains robust ($97.15$\%), performing comparably to \ac{HCL} ($97.21$\%) and HiCervix ($97.29$\%). Because the total number of target classes at the macro level is small, Top-3 classification becomes a less discriminative metric. For the Top-5 accuracy, \ac{SAHC} maintains the overall best performance, outperforming the baseline and the compared methodologies across the evaluated granularities. At the fine-grained level, \ac{SAHC} achieves $68.00$\%, surpassing the HiCervix by $1.31$\%. This performance gap widens at the intermediate level, where \ac{SAHC} reaches $82.66$\% compared to HiCervix $81.08$\%. At the coarse-grained level, \ac{SAHC} achieves $91.88$\%, outperforming the HiCervix by $0.80$\%. Note that Top-5 metrics are omitted at the macro level, as the limited number of classes at this highest taxonomy level renders the Top-5 metric uninformative. The improvement observed in the Top-3 and Top-5 accuracies can be attributed to the design of the \ac{SAHC} architecture. When the network struggles to isolate the exact fine-grained class, the hierarchical consensus mechanism, 
which promotes probabilistic alignment across different hierarchical levels, increases support for semantically related classes within the same hierarchical branch. This reduces the divergence between the projected predictions and the overall committee consensus.

Finally, consistent with the previous analysis, \ac{SAHC} achieves highest mean performance among the compared methods without sacrificing computational efficiency. While architectures like \ac{HRN} and \ac{MGM} impact on the computational efficiency, \ac{SAHC} keeps the computational footprint nearly identical to the baseline model.

\begin{figure*}
  \centering
  \includegraphics[width=\textwidth]{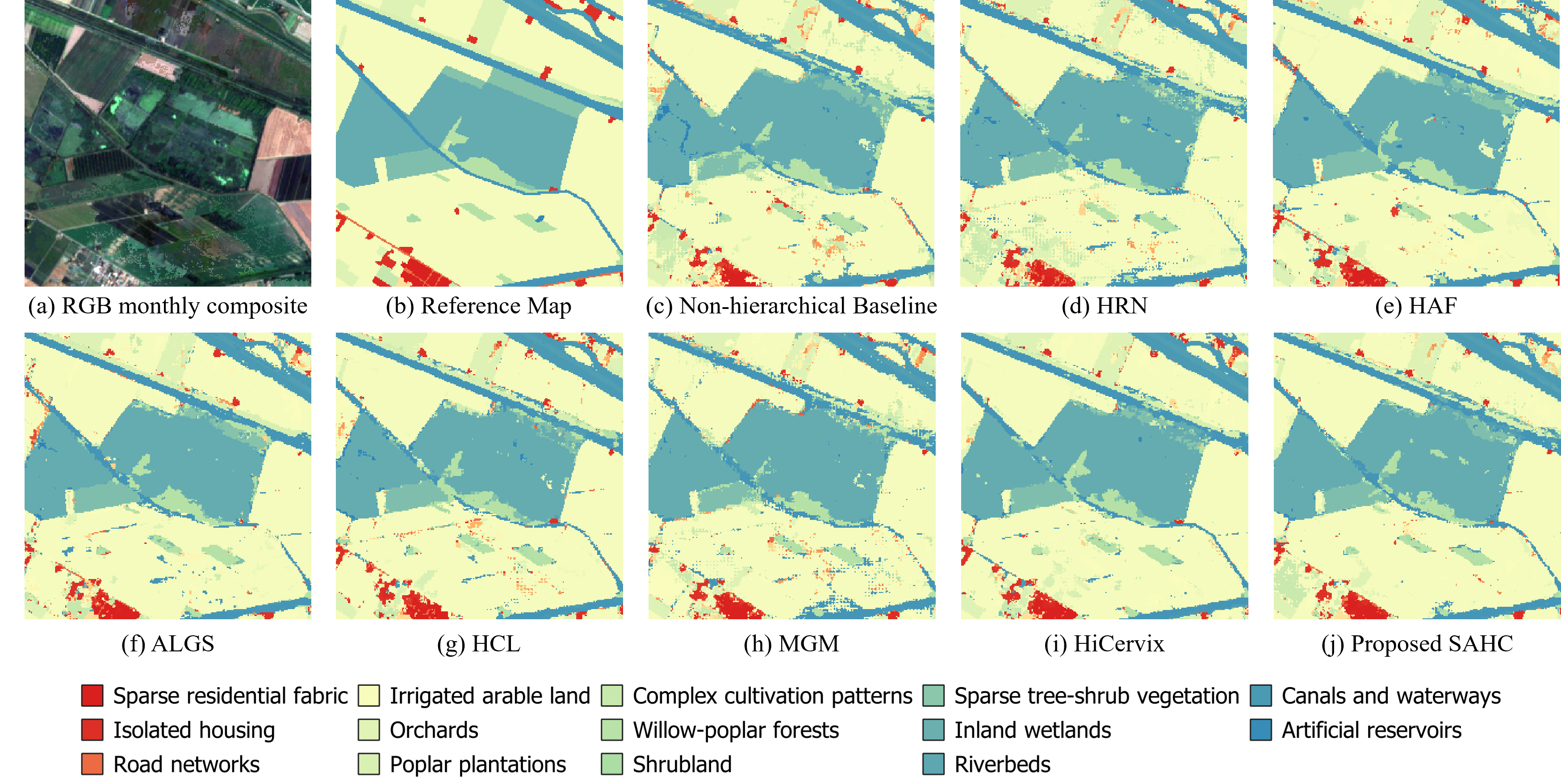}
  \caption{Qualitative example of the predictions at the fine-grained level on the \ac{ELU} dataset obtained by the considered hierarchical methodologies using the Swin Transformer backbone: (a) true color (RGB) monthly composite for August 2020, (b) \ac{CLC} reference map, (c) non-hierarchical baseline (d) HRN method, (e) HAF method, (f) ALGS method, (g) HCL method, (h) MGM method, (i) HiCervix method, and (j) the proposed \ac{SAHC} method.}
  \label{fig:elu_pred_2}
\end{figure*}

Lastly, Fig. \ref{fig:elu_pred} provides a visual comparison of the evaluated hierarchical methodologies on a zoomed portion of the \ac{ELU} dataset. From the qualitative results, one can see that the \ac{SAHC} approach retrieves the most spatially-consistent land-cover maps, characterized by a reduction in classification artifacts compared to the competing approaches. This is particularly evident in the urbanized areas. Indeed, the proposed methodology appears to better characterize the ``isolated housings'', including dominant clusters and smaller patches, while still showing good agreement for the dominant agricultural and woodland classes. These advantages are further evident in Fig. \ref{fig:elu_pred_2}. From a broader perspective, scaling a \ac{RS} classification problem to an extensive taxonomy of $74$ classes inherently introduces severe spectral confusion and extreme class imbalance. In this context, the observed improvements in the \ac{mF1} score are significant. Indeed, they demonstrate that the proposed \ac{SAHC} mitigates the ambiguities typical of such label space, maintaining robust and granular predictions where traditional approaches fail.

\subsection{Sensitivity Analysis of the Consensus Formulation}
The proposed \ac{SAHC} includes two design choices for constructing and regularizing the hierarchical consensus: (i) the divergence measure used in the self-consistent loss, and (ii) the fusion operator used to combine direct and projected predictions. To assess the sensitivity of the method to these choices, we conduct two complementary ablation analyses.
\subsubsection{Sensitivity to the Divergence Measure}
\begin{table}
\centering
\caption{Sensitivity to the Divergence Loss in Consensus. The best results are reported in bold, second best are underlined. \\ The related standard deviation is reported in parentheses}
\label{tab:divergences}
\begin{tblr}{
  colspec = {l *{4}{X[c]}},
  row{1-7} = {fg=black},
  cell{1}{2} = {c},
  cell{1}{3} = {c},
  cell{3}{4} = {Alto,font=\bfseries},
  cell{7}{2,3,5} = {Alto,font=\bfseries},
  cell{1}{2} = {c=2}{},
  cell{1}{4} = {c=2}{},
  hline{1,3,8} = {-}{},
  hline{1,8} = {2}{-}{},
}
 & NWPU & & ELU & \\
Divergence & OA (\%) & mF1 (\%) & OA (\%) & mF1 (\%) \\
MSE & 96.14~(0.05) & 96.14~(0.05) & 38.41~(0.39) & \uline{34.54~(0.39)} \\
KLD & 96.05~(0.10) & 96.05~(0.10) & 38.08~(0.14) & 34.29~(0.17)\\
Sym KLD & 96.21~(0.17) & 96.22~(0.17) & \uline{38.40~(0.30)} & 34.51~(0.22)\\
TV & \uline{96.25~(0.16)} & \uline{96.24~(0.15)} & 38.21~(0.15) & 34.29~(0.16)\\
JSD & 96.55~(0.08) & 96.56~(0.07) & 38.38~(0.31) & 34.56~(0.32)
\end{tblr}
\end{table}
The proposed loss is based on the \ac{JSD} divergence, as the direct and projected predictions are jointly learned distributions, with no fixed reference distribution. We compare it with (i) \ac{KLD}, (ii) symmetric \ac{KLD}, (iii) \ac{MSE}, and (iv) \ac{TV} distance. Table \ref{tab:divergences} shows that \ac{JSD} achieves the best results on NWPU-RESISC45, with $96.55$\% \ac{OA}, and $96.56$\% \ac{mF1}. On \ac{ELU}, the gap tends to shrink. Indeed, \ac{MSE} obtains the highest \ac{OA} ($38.41$\%), whereas \ac{JSD} obtains the highest \ac{mF1} ($34.56$\%), both with a small margin. Overall, the formulation is robust to the selected divergence. We retain \ac{JSD} because it is symmetric, bounded, and remains among the best configurations on both the datasets.
\subsubsection{Sensitivity to the Fusion Operator}
\begin{table}
\centering
\caption{Sensitivity to the Fusion Operator in Consensus. The best \\ results are reported in bold. The related standard\\ deviation is reported in parentheses
}
\label{tab:fusionoperators}
\begin{tblr}{
  colspec = {l *{4}{X[c]}},
  row{1-6} = {fg=black},
  cell{1}{2} = {c=2}{},
  cell{1}{4} = {c=2}{},
  cell{4}{2} = {Alto,font=\bfseries},
  cell{4}{3} = {Alto,font=\bfseries},
  cell{4}{4} = {Alto,font=\bfseries},
  cell{4}{5} = {Alto,font=\bfseries},
  hline{1,3,5} = {-}{},
  hline{1,5} = {2}{-}{},
}
 & NWPU & & ELU & \\
Fusion Operator & OA (\%) & mF1 (\%) & OA (\%) & mF1 (\%)\\
Arithmetic Mean & 96.05~(0.10) & 96.06~(0.08) & 36.31~(0.53) & 32.06~(0.75) \\
Geometric Mean & 96.55~(0.08) & 96.56~(0.07) & 38.38~(0.31) & 34.56~(0.32) 
\end{tblr}
\end{table}
The proposed consensus uses a normalized geometric mean, which favors classes supported by all hierarchical levels. We compare it with arithmetic averaging while keeping all the other components unchanged. As shown in Table \ref{tab:fusionoperators}, geometric fusion outperforms arithmetic fusion on both datasets. On NWPU-RESISC45, it improves \ac{OA} from $96.05$\% to $96.55$\%. On \ac{ELU}, it improves \ac{OA} from $36.31$\% to $38.38$\% and \ac{mF1} from $32.06\%$ to $34.56$\%. Thus, the fusion operator has a larger effect than the specific divergence measure in the tested scenarios, supporting the usage of agreement-sensitive geometric fusion.

\subsection{Sensitivity Analysis to Granularity Weights}
\begin{figure}
    \centering
    \includegraphics[width=\columnwidth]{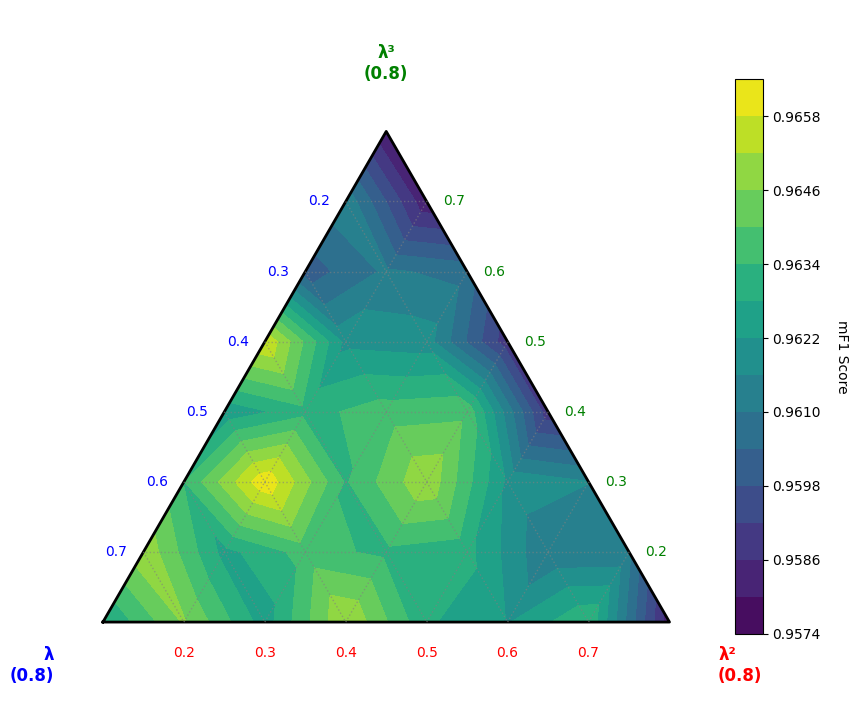}
    \caption{Sensitivity analysis of the hierarchical weights $\lambda^{h}$ on the \ac{VHR} dataset. The performance surface is visualized as a probability simplex, where each point represents a weight combination for the three hierarchical levels sampled at a step size of $0.1$. The simplex heatmap shows the optimal configuration as $\lambda^{1}=0.5$, $\lambda^{2}=0.2$, $\lambda^{3}=0.3$.}
    \label{fig:sensitivity}
\end{figure}
To evaluate the robustness of the proposed hierarchical architecture, we conducted a sensitivity analysis on the \ac{VHR} dataset to assess how model performance varies with the selection of different granularity weights $\lambda^{h}$. The search space was constrained such that the weights sum to one ($\sum_{h=1}^{H}\lambda^{h}=1$), with each $\lambda^{h}$ sampled using a step size of $0.1$. Since the architecture adopts three hierarchical levels, the resulting parameter space can be visualized as a probability simplex that illustrates the various weight combinations across the different granularities. Fig. \ref{fig:sensitivity} illustrates a heatmap reporting the performance across all evaluated configurations of the three layer weights. As demonstrated in the figure, the optimal combination is achieved at $\lambda^{1}=0.5$, $\lambda^{2}=0.2$, $\lambda^{3}=0.3$. This optimal configuration justifies the hypothesis that priority should be allocated to both the fine-grained classification task and the coarsest class. Nevertheless, the analysis also reveals the presence of several local maxima across the performance surface, which are generally skewed toward configurations in favor of the finest-level weight. The best validation performance was obtained by assigning the largest weight to the finest level, a lower weight to the intermediate level, and a moderate weight to the coarse-grained level. The results indicate that the hierarchical levels should emphasize the most discriminative fine-grained objective, while increasing the contribution of the coarsest semantic ones. Intermediate levels are instead down-weighted, providing a better balance between fine-grained discrimination and global semantic regularization.

\subsection{Ablation Study}
\begin{table*}
\centering
\caption{
Ablation study on the ELU and NWPU-RESISC45 datasets. The best results are in bold, second best are underlined}
\label{tab:ablation}
\begin{tblr}{
  row{1-16} = {fg=black},
  colspec = {*{1}{l} *{1}{l} *{1}{l} *{7}{X[c]}},
  hline{1,2,16} = {-}{},
  hline{8} = {dashed},
  hline{1,16} = {2}{-}{},
  cell{2}{1} = {r=6}{}, 
  cell{2,4,6,8,10,12,14}{2} = {r=2}{}, 
  cell{8}{1} = {r=8}{}, 
  cell{2,3,8,9}{9} = {Alto,font=\bfseries}, 
  cell{4,5}{9} = {cmd=\uline}, 
  cell{4,5,6,7,10,11,12,13}{10} = {Alto,font=\bfseries}, 
  cell{2,3,8,9,14,15}{10} = {cmd=\uline}, 
  cell{14,15}{8} = {Alto,font=\bfseries}, 
  cell{6,7,10,11,12,13}{8} = {cmd=\uline}, 
}
Dataset & Level         & Metric    & Baseline      &  +MH           & +MH+{$\hat{\mathbf{p}}_{HC}$}    & +MH+{$\mathcal L_{HC}$}    & SAHC-fixed   & SAHC-direct   & \textbf{SAHC-proj}      \\
NWPU    & Fine          & OA        & 96.12 (0.12)  & 96.26 (0.24)   & 96.30 (0.28)                     & 96.39 (0.20)               & 96.12 (0.05) & 96.59 (0.10)  & 96.55 (0.08)       \\
        &               & mF1       & 96.11 (0.11)  & 96.26 (0.24)   & 96.30 (0.28)                     & 96.39 (0.21)               & 96.13 (0.04) & 96.59 (0.09)  & 96.56 (0.07)       \\
        & Intermediate  & OA        & 97.29 (0.11)  & 97.32 (0.14)   & 97.30 (0.16)                     & 97.42 (0.20)               & 97.28 (0.12) & 97.47 (0.08)  & 97.48 (0.07)       \\
        &               & mF1       & 96.59 (0.10)  & 96.25 (0.21)   & 96.48 (0.21)                     & 96.65 (0.30)               & 96.49 (0.16) & 96.72 (0.13)  & 96.75 (0.16)       \\
        & Coarse        & OA        & 97.43 (0.08)  & 97.65 (0.33)   & 97.79 (0.16)                     & 97.79 (0.26)               & 97.80 (0.10) & 97.42 (0.69)  & 97.90 (0.07)       \\
        &               & mF1       & 97.07 (0.10)  & 97.35 (0.25)   & 97.43 (0.17)                     & 97.42 (0.28)               & 97.45 (0.10) & 97.54 (0.13)  & 97.58 (0.09)       \\
ELU     & Fine          & OA        & 36.84 (0.37)  & 38.03 (0.50)   & 37.73 (0.41)                     & 38.29 (0.18)               & 38.15 (0.25) & 38.39 (0.36)  & 38.38 (0.31)       \\
        &               & mF1       & 32.57 (0.23)  & 33.89 (0.41)   & 33.57 (0.32)                     & 33.96 (0.15)               & 34.35 (0.27) & 34.67 (0.34)  & 34.56 (0.32)       \\
        & Intermediate  & OA        & 47.84 (0.46)  & 48.64 (0.51)   & 49.02 (0.56)                     & 48.93 (0.31)               & 49.79 (0.33) & 49.13 (0.53)  & 50.20 (0.13)       \\
        &               & mF1       & 40.64 (0.38)  & 41.31 (0.41)   & 41.59 (0.54)                     & 41.27 (0.33)               & 42.61 (0.47) & 42.22 (0.23)  & 42.97 (0.37)       \\
        & Coarse        & OA        & 57.64 (0.07)  & 58.63 (0.48)   & 59.29 (0.34)                     & 58.92 (0.52)               & 59.98 (0.21) & 58.97 (0.50)  & 60.21 (0.21)       \\
        &               & mF1       & 52.89 (0.14)  & 54.14 (0.36)   & 54.66 (0.21)                     & 54.46 (0.30)               & 55.39 (0.06) & 54.56 (0.55)  & 55.54 (0.15)       \\
        & Macro         & OA        & 77.55 (0.04)  & 77.58 (0.08)   & 77.89 (0.18)                     & 77.91 (0.19)               & 78.46 (0.12) & 77.80 (0.16)  & 78.33 (0.17)       \\
        &               & mF1       & 68.35 (0.18)  & 68.29 (0.08)   & 68.72 (0.17)                     & 68.10 (0.27)               & 69.10 (0.43) & 68.33 (0.22)  & 69.06 (0.16)       \\
\end{tblr}
\end{table*}
The performance contribution of each component in the proposed hierarchical approach is analyzed in the ablation study on both datasets presented in Table \ref{tab:ablation}. In this study, we evaluate the ResNet50 backbone on NWPU-RESISC45 and the Swin Transformer backbone on \ac{ELU}. The considered configurations include the following cases:
\begin{enumerate}
    \item \textbf{Baseline}: the standard approach with no multi-level classifiers;
    \item \textbf{Multi-Heads (MH)}: the baseline approach modified with additional classification heads for the coarse-grained and intermediate classes.
    \item \textbf{Multi-level with \ac{HC} inference} $\hat{\mathbf{p}}_{HC}$: the multi-level approach with consensus applied solely during inference (considering frozen \textit{user-defined} hierarchies).
    \item \textbf{Multi-level with $\mathcal{L}_{HC}$ training}: the multi-level approach trained with the self-consistent loss, excluding consensus-based classification during inference.
    \item \textbf{SAHC with fixed projectors (SAHC-fixed)}: a taxonomy-guided \ac{SAHC} in which the projectors are obtained by directionally normalizing the \textit{user-defined} hierarchy indicator matrices and are kept frozen throughout training. 
    \item \textbf{\ac{SAHC} with direct-head inference (SAHC-direct)}: the adaptive-projector \ac{SAHC} configuration evaluated using the direct hierarchy-specific heads at inference.
    \item \textbf{\ac{SAHC} with adaptive projectors (SAHC-proj)}: the adaptive-projector \ac{SAHC} configuration, which jointly optimizes taxonomy-initialized hierarchy projections together with the remaining \ac{SAHC} components, consensus inference, and the supervised consensus loss ($\mathcal{L}^{HC}_{CCE}$) at multiple levels.
\end{enumerate}

Table \ref{tab:ablation} reports the contribution of the different modules composing \ac{SAHC}. Adding multi-head supervision alone improves the fine-level performance on both datasets, with a modest gain on NWPU-RESISC45 and a larger improvement on \ac{ELU}. In particular, fine-level \ac{mF1} on \ac{ELU} increases from $32.57$\% to $33.89$\%, indicating that multi-level supervision is beneficial in the deeper and more complex hierarchies. The consensus-training objectives further improve the direct-head model. The Baseline + MH + {$\hat{\mathbf{p}}_{HC}$} case shows the effect of fixed taxonomy-based aggregation alone. On NWPU-RESISC45, its effect is limited. On \ac{ELU}, fixed consensus slightly reduces fine-level performance relative to the multi-head model, while improving the intermediate, coarse, and macro-level results. This confirms that fixed aggregation can favor cross-level agreement. Compared with Baseline + MH +{$\mathcal L_{HC}$} case, \ac{SAHC}-direct increases fine-level \ac{mF1} by $0.20$\% on NWPU-RESISC45 and by $0.71$\% on \ac{ELU}. Applying consensus inference after complete training leaves fine-level performance nearly unchanged, while improving the intermediate, coarse, and macro-level \ac{ELU} results. Thus, consensus inference mainly benefits coherent multi-level outputs and coarser-level performance rather than the fine-level score alone. The preferred inference output is application-dependent. Indeed, $\hat{\mathbf{p}}^{h}_{\mathrm{HC}}$ is preferable when coherent predictions are required across multiple hierarchical levels or when high performances in coarser label predictions are favored, whereas $\hat{\mathbf{p}}^{h}$ can be retained when only the finest-level metric is the operational objective.

Finally, the comparison between \ac{SAHC}-fixed and the adaptive-projector \ac{SAHC} configuration (SAHC-proj) isolates the calibration provided by the trainable hierarchy projectors. Adaptive projectors improve fine-level \ac{OA} and \ac{mF1} on NWPU-RESISC45 by approximately $0.43$\%. On \ac{ELU}, they provide smaller gains at the fine, intermediate, and coarse levels, while macro-level performance remains comparable. Overall, the ablation indicates that the main benefit of \ac{SAHC} arises from multi-level supervision and self-consistent training. Consensus inference further supports coherent predictions across hierarchical levels, whereas adaptive projectors provide a modest, level-dependent calibration beyond frozen taxonomy-derived projectors.

\subsection{Sensitivity to Projectors Initialization}
\begin{table}
\caption{
Sensitivity to the projector's initialization. The best results\\are reported in bold, second best are underlined. The\\related standard deviation is reported in parentheses}
\label{tab:sensitivity_inits}
\begin{tblr}{
  row{1-16} = {fg=black},
  colsep = 1pt, 
  colspec = {*{1}{l} *{1}{l} *{1}{l} *{4}{X[c]}},
  hline{1,2,16} = {-}{},
  hline{8} = {dashed},
  hline{1,16} = {2}{-}{},
  cell{2}{1} = {r=6}{}, 
  cell{2,4,6,8,10,12,14}{2} = {r=2}{}, 
  cell{8}{1} = {r=8}{}, 
  cell{8,9}{4} = {Alto,font=\bfseries}, 
  cell{14}{4} = {cmd=\uline}, 
  cell{13}{5} = {Alto,font=\bfseries}, 
  cell{8,9}{5} = {cmd=\uline}, 
  cell{14,15}{6} = {Alto,font=\bfseries}, 
  cell{2,3,4,5,6,7,10,11,12,13}{6} = {cmd=\uline}, 
  cell{2,3,4,5,6,7,10,11,12}{7} = {Alto,font=\bfseries}, 
  cell{15}{7} = {cmd=\uline}, 
}
Dataset & Level         & Metric    & $\Delta=0$      &  $\Delta=1$      &  $\Delta=3$     & $\Delta=5$     \\
NWPU    & Fine          & OA        & 96.24 (0.05)    &  96.17 (0.04)    &  96.34 (0.11)   & 96.55 (0.08)   \\
        &               & mF1       & 96.24 (0.05)    &  96.18 (0.04)    &  96.35 (0.10)   & 96.56 (0.07)   \\
        & Inter         & OA        & 97.33 (0.02)    &  97.21 (0.06)    &  97.44 (0.05)   & 97.48 (0.07)   \\
        &               & mF1       & 96.55 (0.02)    &  96.49 (0.07)    &  96.70 (0.07)   & 96.75 (0.16)   \\
        & Coarse        & OA        & 97.80 (0.02)    &  97.67 (0.11)    &  97.85 (0.15)   & 97.90 (0.07)   \\
        &               & mF1       & 97.45 (0.02)    &  97.35 (0.15)    &  97.51 (0.16)   & 97.58 (0.09)   \\
ELU     & Fine          & OA        & 38.76 (0.17)    &  38.54 (0.36)    &  38.48 (0.31)   & 38.38 (0.31)   \\
        &               & mF1       & 34.96 (0.15)    &  34.64 (0.43)    &  34.56 (0.28)   & 34.56 (0.32)   \\
        & Inter         & OA        & 49.72 (0.28)    &  50.02 (0.41)    &  50.11 (0.31)   & 50.20 (0.13)   \\
        &               & mF1       & 42.61 (0.21)    &  42.81 (0.56)    &  42.82 (0.47)   & 42.97 (0.37)   \\
        & Coarse        & OA        & 59.76 (0.06)    &  60.08 (0.19)    &  60.20 (0.23)   & 60.21 (0.21)   \\
        &               & mF1       & 55.47 (0.12)    &  55.59 (0.17)    &  55.55 (0.25)   & 55.54 (0.15)   \\
        & Macro         & OA        & 78.39 (0.13)    &  78.28 (0.11)    &  78.43 (0.14)   & 78.33 (0.17)   \\
        &               & mF1       & 69.04 (0.29)    &  69.02 (0.22)    &  69.11 (0.26)   & 69.06 (0.16)   \\
\end{tblr}
\end{table}
The hierarchy parameters in \eqref{eq:wfromindicator} are initialized in the unconstrained logit space from the binary indicator matrices. We evaluate different values of $\Delta$, which controls the logit gap between valid and invalid hierarchical connections. The case $\Delta=0$ corresponds to a uniform initialization, whereas larger values of $\Delta$ impose a stronger taxonomy prior while preserving the trainability of all projectors. 
\begin{figure*}
  \centering
  \includegraphics[width=\textwidth]{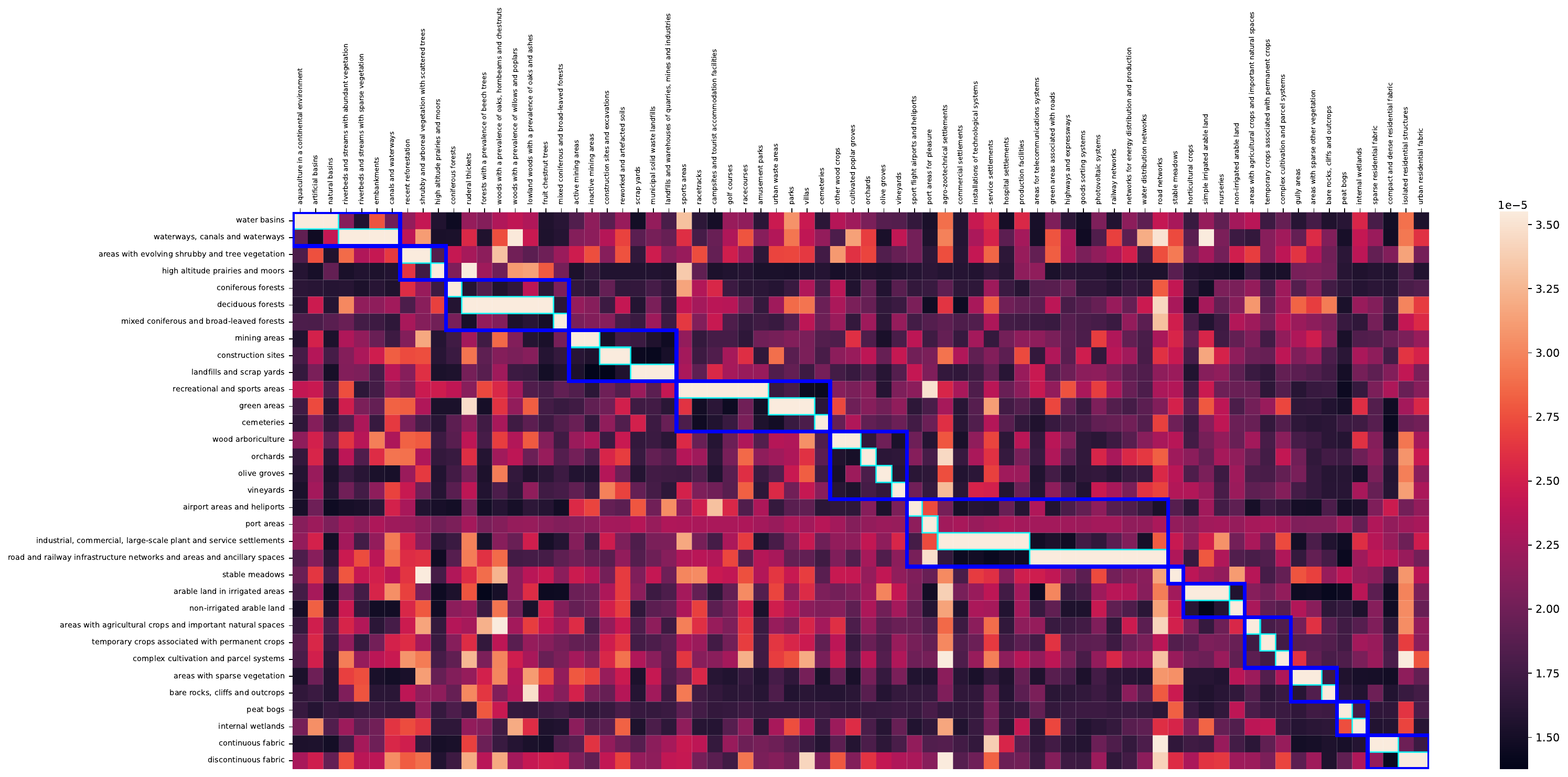}
  \caption{Heatmap of the estimated joint probability matrix $\mathbf{J}^{h_{2},h_{1}}$ considering the Swin Transformer backbone on the \ac{ELU} dataset from intermediate classes to fine-grained classes. The cyan rectangles identify the expected hierarchical aggregation of the fine-grained classes at the intermediate level.}
  \label{fig:heatmap}
\end{figure*}

As reported in Table \ref{tab:sensitivity_inits}, the effect of $\Delta$ is limited and varies across datasets and hierarchical levels. On NWPU-RESISC45, larger values provide slightly better average results, whereas on \ac{ELU} smaller values are marginally preferable at the finest level and larger values are marginally preferable at some coarser levels. This trend may suggest that stronger priors are more effective where semantic aggregation is more stable, while finer-grained classes may benefit from a more flexible calibration of cross-level relations. However, these differences are negligible, as suggested by the observed standard deviations over three runs. Overall, the results indicate that \ac{SAHC} is only moderately sensitive to $\Delta$, and we retain $\Delta=5$ as the common configuration adopted in the main experiments on both datasets.

To further assess the stability of the learned hierarchy matrices, we compared the final projector matrices after training with different random seeds.
With the default strong-prior initialization (\textit{i.e.}, $\Delta=5, s=0.01$), the average cosine similarity and Pearson correlation between learned matrices were both $99.81\%$. When increasing the initialization noise to $s=0.2$, the corresponding similarities remained high, namely $97.52\%$ and $97.45\%$, respectively. These results indicate that the learned semantic relationships are stable across training runs and are not artifacts of a particular random initialization.

\subsection{Hierarchy Projection Matrices Analysis}
To characterize the cross-level relations encoded by the adaptive projectors we analyze the estimated joint probability matrix $\mathbf{J}^{h_{2},h_{1}}$ representing adaptive relationships between the intermediate- and fine-grained levels
for the \ac{ELU} dataset. The corresponding visualization on the NWPU dataset is omitted, since the learned projectors do not reveal qualitatively new off-taxonomy structures. 
Fig. \ref{fig:heatmap} presents a heatmap of this joint probability matrix between intermediate- and fine- level classes, illustrating the hierarchical structure captured by the proposed \ac{SAHC} methodology on the \ac{ELU} dataset. The visualization confirms that the adaptive projectors remains anchored to the \textit{user-defined} semantic aggregations between the levels. Specifically, the cyan rectangles in the heatmap correspond to the \textit{user-defined} expected mappings, from fine-grained classes to intermediate classes. The higher weights within these regions confirm that the \textit{data-driven} estimated mappings largely preserve the intended \textit{user-defined} taxonomy. In addition to these rigid hierarchies, the network also captures secondary cross-level relationships. Intuitively, one can expect that these secondary cross-level relationships correspond to relationships related to coarser hiearchical levels. To help visualize this, Fig. \ref{fig:heatmap} shows with blue rectangles how the classes at the fine level aggregate at the coarse level. However, because separate joint matrices are learned for different hierarchical transitions in the proposed architecture, the model is able to assign secondary relationships weights that largely ignore the initial \textit{user-defined} hierarchy.
For instance, the adaptive weights indicate active associations between ``areas with agricultural crops and important natural spaces'' and ``woods with a prevalence of oaks, hornbeams and chestnuts", as well as between ``continuous fabric" and ``road networks". These weights provide a qualitative example of recurrent cross-level ambiguity in the dataset, highlighting the importance of a \textit{data-driven} calibration. Nonetheless, some associations with relatively high weights show semantically wrong relationships, such as ``high altitude prairies and moors'' and ``sport areas'' or ``airports'' and ``campsites'', which could be related to both spectral similarity, and semantic ambiguity.

Finally, we examined the dominant cross-level weights within the adaptive hierarchy projectors to interpret the task-dependent associations used by the projectors. For the adjacent forward projectors, we ranked the learned weights corresponding to the hierarchical transitions and extracted the five most significant relations, reported in Table \ref{tab:foundtransitions}. At finer levels, several dominant weights correspond to plausible spatial co-occurrences, such as ``road networks'' with ``continuous fabric'' or ``waterways, and canals'' with ``simple irrigated arable land''. Toward the coarsest levels, pairs such as ``urbanized areas'' and ``agricultural territories'' are more likely attributable to spectral similarities, comparable structures, or classifier confusions rather than newly discovered semantic hierarchy links.

\begin{table}
\centering
\caption{Top-5 cross-level semantic transitions extracted from the learned forward hierarchy projectors. The table reports \\ the associations with the highest transition weights \\ mapped from the coarser hierarchy level $h$ to \\ the finer level $\hat{h}$}
\label{tab:foundtransitions}
\begin{tblr}{
  colspec = {l *{2}{X[l,m]}},
  row{1-17} = {fg=black},
  hline{3-16} = {dashed},
  cell{2,7,12}{1} = {r=5}{},
  row{1} = {c,font=\bfseries},
  hline{1-2,7,12,17} = {-}{},
  hline{1,17} = {2}{-}{},
}
$\mathbf{J}^{h,\hat{h}}$    & $h$                                                                           & $\hat{h}$ \\
$\mathbf{J}^{h_2,h_1}$  & continuous fabric & road networks                                                            \\
                        & discontinuous fabric & agro-zootechnical settlements                                            \\
                        & high altitude prairies and moors & ruderal thickets                                                         \\
                        & areas with agricultural crops and important natural spaces & woods with a prevalence of oaks, hornbeams and chestnuts                 \\
                        & waterways, and canals & simple irrigated arable land                                             \\
$\mathbf{J}^{h_3,h_2}$  & urbanized areas & road and railway infrastructure networks and areas and ancillary spaces  \\
                        & urbanized areas & industrial, commercial, large-scale plant and service settlements        \\
                        & heterogeneous agricultural areas & arable land in irrigated areas                                           \\
                        & maritime wetlands & arable land in irrigated areas                                           \\
                        & non-irrigated arable land & stable meadows                                                           \\
$\mathbf{J}^{h_4,h_3}$  & artificially modeled territories & arable land                                                              \\
                        & artificially modeled territories & heterogeneous agricultural areas                                         \\
                        & agricultural territories & environments with evolving shrubby and/or herbaceous vegetation          \\
                        & wooded territories and semi-natural environments & heterogeneous agricultural areas                                         \\
                        & urbanized areas & agricultural territories                                                 \\
\end{tblr}
\end{table}

\subsection{Discussion}
The analysis of the results reveal that the proposed methodology is able to exploit both complex and simple hierarchies. Indeed, \ac{SAHC} achieves the most accurate overall results, obtaining the best fine-, intermediate-, and coarse- level performance on \ac{ELU} and the best results across all levels on NWPU-RESISC45, while remaining competitive at the \ac{ELU} macro level.
Furthermore, the self-consistency loss effectively constrained the hierarchical levels to converge toward a unified prediction, promoting cross-level agreement in the final classification maps. The experiments demonstrate that SAHC provides the best overall trade-off between fine-grained discrimination and coherent multi-level prediction. The ablation further shows that the learned self-consistency is the principal contributor to fine-level performance, whereas consensus inference primarily improves intermediate, coarse, and macro-level predictions. Therefore, direct-head predictions may be preferable when only the finest-level discriminative metric is operationally relevant, while consensus outputs are more appropriate when the downstream application requires coherent predictions across the hierarchy or better performance at coarser levels.

The proposed SAHC formulation is lightweight for the three- and four- level taxonomies considered in this study. Its dense cross-level parameterization nevertheless scales quadratically with hierarchy depth and with the products of the involved label cardinalities. 
Taxonomies substantially deeper than those evaluated here may therefore benefit from adjacent-level, sparse, or selectively connected consensus variants.

A limitation of the proposed method is its reliance on a predefined hierarchy to initialize and train the hierarchical consensus. While the hierarchy matrices are trainable and allow \textit{data-driven} calibration of the projection weights,
the present study does not explicitly evaluate robustness to intentionally corrupted, incomplete, or partially inaccurate hierarchy definitions. If the predefined hierarchy poorly reflects the semantic relationships present in the data, the initial hierarchy prior and the supervised losses at the coarser levels may bias the learned projection matrices toward suboptimal inter-level compatibility.
Furthermore, this constraint bounds transferability to geographically distinct regions or different \ac{LC} distributions with misaligned or structurally heterogeneous taxonomies.

\section{Conclusion}\label{conclusion}

In scene classification and semantic segmentation \ac{RS} tasks, the hierarchical relationship among the target classes is often under-exploited, resulting in the exclusion of valuable semantic information from the decision-making process. To address this limitation, this paper has proposed the \ac{SAHC} framework, which combines hierarchy-specific classification branches, cross-level projections, geometric consensus, self-consistent training, supervised consensus learning, and optional consensus inference. Taxonomy-derived projectors provide the core of the method, while trainable hierarchy log-joint matrices offer an adaptive tuning for global cross-level relationships around the \textit{user-defined} semantic prior.
The proposed methodology is lightweight and scalable, making it adaptable to different architectures and datasets organized in a hierarchical manner. The experimental results demonstrated the effectiveness of the proposed approach on two different \ac{RS} datasets, related to the classification of \ac{VHR} and \ac{MS} images. Indeed, the proposed approach is particularly relevant in \ac{VHR} image and \ac{MS} \ac{SITS} classification, where the semantic structure of the data and available datasets support an explicit hierarchical representation of the task. 

Future work will investigate robustness under hierarchy perturbations, assessing the ability of the adaptive projectors to compensate for incomplete or partially inaccurate taxonomy definitions. Moreover, adaptation across wide geographic areas will be explored under misaligned or structurally heterogeneous taxonomies, evaluating the transferability and recalibration of the learned cross-level relationships.

\bibliographystyle{IEEEtran}
\bibliography{references}

\begin{IEEEbiography}[{\includegraphics[width=1in,height=1.25in,clip]{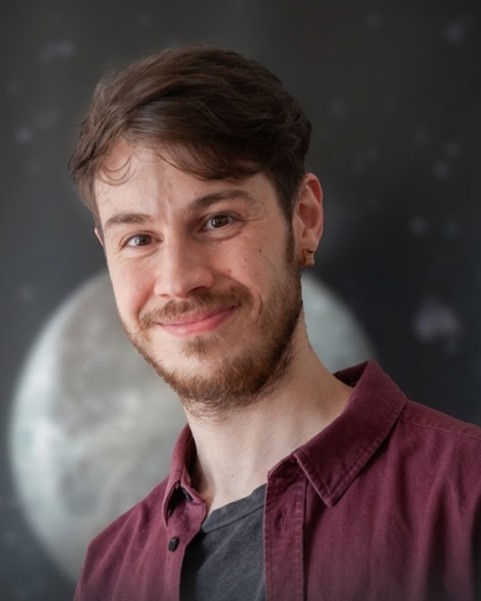}}]{\textbf{Giulio Weikmann}} (Member, IEEE) received the M.S. degree (summa cum laude) in information and communication engineering from the University of Trento, Trento, Italy, in 2020, and the Ph.D. degree (cum laude) in information engineering and computer science from the University of Trento in 2024.

He is currently an Assistant Professor with the Remote Sensing Laboratory, Department of Information Engineering and Computer Science, University of Trento, Trento, Italy. His research interests include deep learning methods for the automatic processing and classification of remotely sensed optical data, with particular emphasis on multitemporal data and time-series characterization and analysis. He has participated in several national and international research projects and has coordinated his research team's contributions to the European Commission H2020 projects \textit{ExtremeEarth} and \textit{AI4Copernicus}, and the Italian Space Agency (ASI) projects \textit{PrismaLearn} and \textit{Space It Up!}.

Dr. Weikmann was a recipient of the 17th Eugenio Zilioli Award for the Best Master's Thesis from the Associazione Italiana Telerilevamento and the Consiglio Nazionale delle Ricerche, and the 2023 Best Student Paper Award at the \textit{SPIE Image and Signal Processing for Remote Sensing XXIX Conference}. He has served as a reviewer for the \textsc{IEEE Transactions on Geoscience and Remote Sensing} and the \textsc{IEEE Geoscience and Remote Sensing Letters}.
\end{IEEEbiography}

\vspace*{-2.2\baselineskip}
\begin{IEEEbiography}[{\includegraphics[width=1in,height=1.25in,clip]{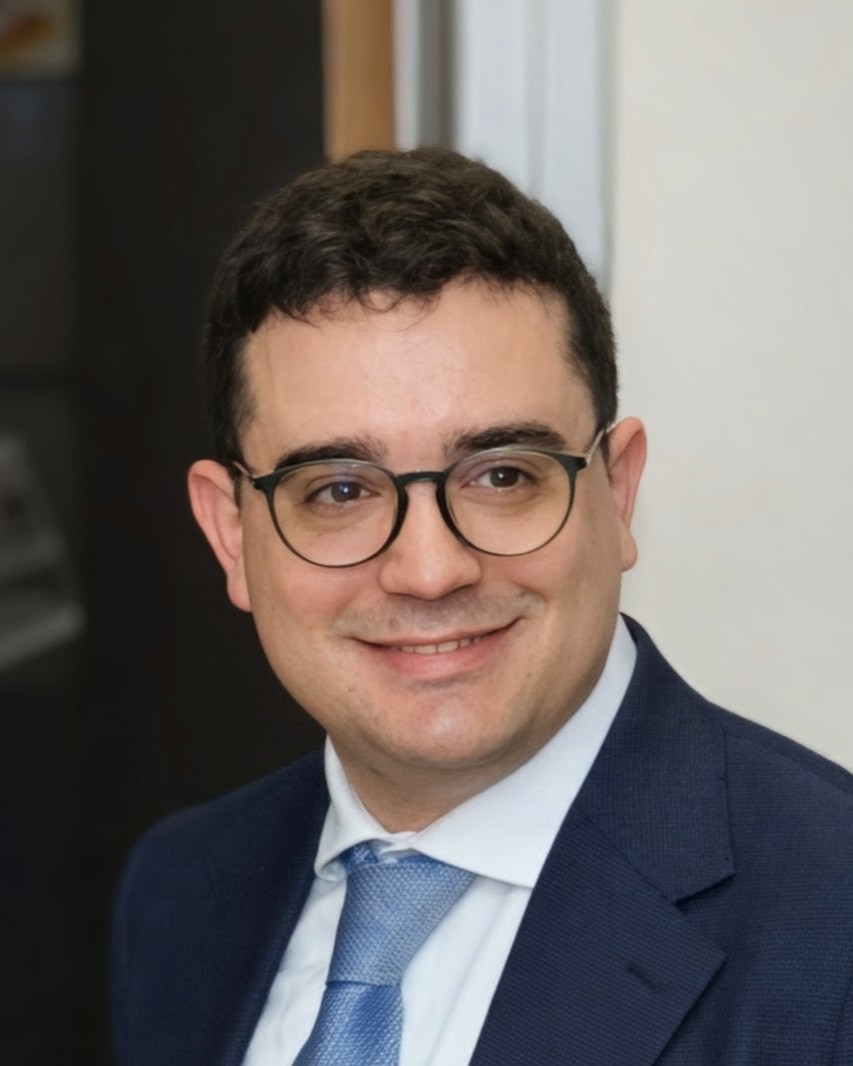}}]{\textbf{Gianmarco Perantoni}} (Member, IEEE)
received the M.S. degree (summa cum laude) in information and communication engineering from the University of Trento, Trento, Italy, in 2020, and the Ph.D. degree (cum laude) in information engineering and computer science from the University of Trento in 2025. 

He is currently a Postdoctoral Researcher and Teaching Assistant with the Remote Sensing Laboratory (https://rslab.disi.unitn.it/) in the Department of Information Engineering and Computer Science, University of Trento, Trento, Italy. His research interests include deep learning for the automatic processing and classification of remote-sensed optical data, with a specific focus on land cover mapping using weak supervision. His current work supports the \textit{High Resolution Land Cover} project in the framework of the \textit{Climate Change Initiative} of the European Space Agency (ESA).

Dr. Perantoni was the recipient of the 2021 IEEE GRS29-CNI-SI Prize for the three best Italian master's theses in geoscience and remote sensing. He serves as a reviewer for the \textsc{IEEE Transactions on Geoscience and Remote Sensing}, the \textsc{IEEE Journal of Selected Topics in Applied Earth Observations and Remote Sensing}, the \textsc{IEEE Geoscience and Remote Sensing Letters}, and the IEEE International Geoscience and Remote Sensing Symposium (IGARSS).
\end{IEEEbiography}
\vfill

\begin{IEEEbiography}[{\includegraphics[width=1in,height=1.25in,clip]{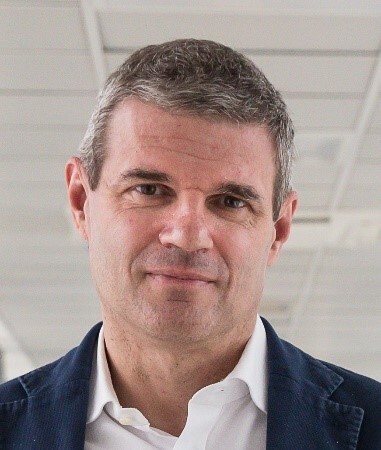}}]{Lorenzo Bruzzone} (Fellow, IEEE) received the Laurea (M.S.) degree in electronic engineering (summa cum laude) and the Ph.D. degree in telecommunications from the University of Genoa, Italy, in 1993 and 1998, respectively. He is currently a Full Professor of telecommunications at the University of Trento, Italy, where he teaches remote sensing, radar, and digital communications.

Dr. Bruzzone is the founder and the director of the Remote Sensing Laboratory (https://rslab.disi.unitn.it/) in the Department of Information Engineering and Computer Science, University of Trento. His current research interests are in the areas of remote sensing, radar and SAR, signal processing, machine learning and pattern recognition. He promotes and supervises research on these topics within the frameworks of many national and international projects related to Earth observation and planetary exploration. He is the Principal Investigator of many research projects. Among the others, he is currently the Principal Investigator of the \textit{Radar for Icy Moon Exploration (RIME)} and of the \textit{Subsurface Radar Sounder (SRS)} instruments in the framework of the JUICE and the Envision missions, respectively, of the European Space Agency (ESA) and is the Science Lead for the \textit{High Resolution Land Cover} project in the framework of the \textit{Climate Change Initiative} of ESA. He is the author (or coauthor) of more than 450 scientific publications in referred international journals, more than 410 papers in conference proceedings, and 25 book chapters. He is editor/co-editor of 18 books/conference proceedings and 2 scientific books. His papers are highly cited, as proven from the total number of citations (more than 65000) and the value of the h-index (119) (source: Google Scholar). He was invited as keynote speaker at more than 40 international conferences and workshops. He was a member of the Administrative Committee of the IEEE Geoscience and Remote Sensing Society (GRSS) between 2009 and 2023, where since 2019 he was Vice-President for Professional Activities. Dr. Bruzzone ranked first place in the Student Prize Paper Competition of the 1998 IEEE International Geoscience and Remote Sensing Symposium (IGARSS), Seattle, July 1998. Since that he was recipient of many international and national honors and awards, including the recent \textit{IEEE GRSS 2015 Outstanding Service Award}, the 2017 and 2018 \textit{IEEE IGARSS Symposium Prize Paper Awards}, the \textit{2019 WHISPER Outstanding Paper Award} and the \textit{2022 Letter Prize Paper Award} for the best paper published on the \textsc{IEEE Geoscience and Remote Sensing Letters}. Dr. Bruzzone was a Guest Co-Editor of many Special Issues of international journals. He is the co-founder of the IEEE International Workshop on the Analysis of Multi-Temporal Remote-Sensing Images (MultiTemp) series and is currently a member of the Permanent Steering Committee of this series of workshops. Since 2003 he has been the Chair of the SPIE Conference on Image and Signal Processing for Remote Sensing. He has been the founder of the \textsc{IEEE Geoscience and Remote Sensing Magazine} for which he has been Editor-in-Chief between 2013-2017. He was \textit{Distinguished Speaker} of the IEEE Geoscience and Remote Sensing Society between 2012-2016.

\end{IEEEbiography}%

\end{document}